\documentclass[12pt,letterpaper]{report}

\RequirePackage{setspace}
\usepackage{boxedminipage}

\usepackage[titles]{tocloft} 
\cftsetindents{figure}{0em}{3.5em}
\cftsetindents{table}{0em}{3.5em}

\usepackage{ifpdf} 

\newif\iftestbox

\RequirePackage{TUthesis}
\usepackage{amsmath} 
\usepackage{amssymb}  
\usepackage{graphicx}
\usepackage{hyperref}
\usepackage{pgfplots}
\pgfplotsset{width=10cm,compat=1.9}
\usepackage{todonotes}

\begin{document}

\testboxfalse 
\iftestbox \testboxex \fi

%
%
\titleoneline{CramNet: Layer-wise neural network compression with a teacher}
\title{CramNet: Layer-wise Deep Neural Network \\ Compression with Knowledge Transfer \\ from a Teacher Network}  
\author{Jon Hoffman}
\degreename{Master of Science}
\dept{Computer Science}
\submityear{2018}

%
%
\coadvisorfalse  

%
%
\committeesize=3

\advisor{Sandip Sen}    
\secondmember{Kaveh Ashenayi} 
\thirdmember{John Hale}   
\fourthmember{Fourth Member} 
\fifthmember{Fifth Member}   
\sixthmember{Sixth Member}   

\numofpages{47}                    
\numofchapters=5                   
\lastchapter{Conclusions and Future Work}	  
\numofabstractwords{98}            

%
%
\thesistrue  

%
%

\copyrighttrue     
\figurespagetrue   
\tablespagetrue    

\beforeabstract    
\abstractp         

%
%
Neural Networks accomplish amazing things, but they suffer from computational and memory bottlenecks that restrict their usage.  Nowhere can this be better seen than in the mobile space, where specialized hardware is being created just to satisfy the demand for neural networks.  Previous studies have shown that neural networks have vastly more connections than they actually need to do their work.   This thesis develops a method that can compress networks to less than 10\% of memory and less than 25\% of computational power, without loss of accuracy, and without creating sparse networks that require special code to run.

\acknowledgementsp
%
%
I would like to thank everyone, especially Melanie and my parents who supported me.  I also thank my employer, L3 Aeromet, which paid for my education.  Last, but not least my professors, without whom I would know less, and my life would be poorer.

\afteracknowledgementsp


\TUchapter{Introduction}
Deep neural networks (Deep NNs) represent the state-of-the-art for many computer vision tasks \cite{AlexNet}~\cite{GoogleNet}.  Over the past decade or so, Deep NNs have emerged from the confines of university labs where they were a topic primarily of interest to only a small group of academic researchers, to become what is perhaps the most widely celebrated and used machine learning technique.  Deep NNs are now the primary focus of a large swatch of application developers working with large, real-life data sets in the industry and academia alike.  In 2015, a research team from Microsoft surpassed human accuracy for the image classification task~\cite{Better_than_human}, and progress has not abated since.

Today, neural networks are used in Google Photos to identify image contents, in Google Translate to improve comprehension and perform real-time audio translation, and even Microsoft Powerpoint~\cite{ppt} to provide real-time captions. While the capabilities of modern networks are very impressive, their sheer size can limit their effectiveness.  Networks like AlexNet and VGG-16 contain millions or hundreds of millions of parameters, requiring megabytes of memory and teraFLOPs of processing power.  

The significant memory usage may not be a concern in desktop or laboratory applications, but in the mobile space or on cheaper computers, where bandwidth, memory, and processing power are limited, the size of a neural network can prove to be prohibitively costly.  The inability to run these networks where they are needed has, in part, driven a rise in cloud computing~\cite{cloud}.  Some mobile phones now contain specialized chips just to run neural networks on the device~\cite{chip}.

In almost all cases, the speed of operation of the network is also of concern.  Much effort has gone into improving the speed at which neural networks execute, from dedicated libraries, to Graphics Processing Units (GPUs), to Tensor Processing Units (TPUs)~\cite{TPU}.  In all these cases, smaller networks have demonstrated the potential for faster execution.

Several methods have been developed to address both memory usage and speed.  The simplest methods such as AlexNet~\cite{AlexNet} and SqueezeNet~\cite{SqueezeNet} involve creating new architectures.  Since these architectures often have lower accuracy than the larger versions they were to replace, additional methods such as Optimal Brain Surgeon~\cite{OBS} and Deep Compression~\cite{Deep-Compression} have been created to directly address the size problem.  Unfortunately, these early pruning methods all created sparse networks, which either reduced speed or required special hardware. Hence a new generation of compression methods such as FitNets~\cite{FitNets} and ThiNet~\cite{ThiNet} were developed to at least partially address the issue.

CramNet was created to use the best of all the methods.  Unlike Deep Compression or OBS, it produces dense networks.  Like FitNets, a teacher network helps preserve learned accuracy.  And like ThiNet, it operates in a layer-wise manner to assist in parameter exploration.  Unlike all of these methods, it supports completely different architectures.  This thesis will compare the CramNet compression method with the current state-of-the-art on two different image classification datasets, demonstrating competitive results.

\TUchapter{Related Work}

This chapter explains the basis of neural networks (NNs), and describes several previous methods that are among those compared with the proposed CramNet.  In addition, it describes the technical justification for why NN compression is necessary and specifically how current methods work and where they fall short.


\TUsection{Layered Feed-forward NNs}

Layered, feed-forward NNs have been extensively used in diverse applications since the development of the backpropagation method to train NNs~\cite{backpropagation}.  Backpropagation makes it possible to automatically train the weights and biases of a neural network given a labeled dataset:
\begin{equation}
    \label{eq:backprop}
    \Delta w_{ij}=-\eta \frac{\partial E}{\partial w_{ij}}.
\end{equation}
is the change in a network weight \(w_{ij}\), given the learning rate \(\eta\), and the portion of the error due to the weight.  This equation is the foundation of most modern machine learning.

Each layer has a set of inputs \(P\), and a set of neurons \(S\).  In a typical fully connected layer, the output of each neuron is 
\begin{equation}
    \label{eq:neuron}
    S_i = \sum_{p\in P}{w_{ip}*p} + b_i,
\end{equation}
where \(w_{ip}\) is the weight connecting input \(p\) with neuron \(i\), and \(b_i\) is the bias of neuron \(i\).

\begin{figure}[ht]
    \centering
    \includegraphics[width=.85\linewidth]{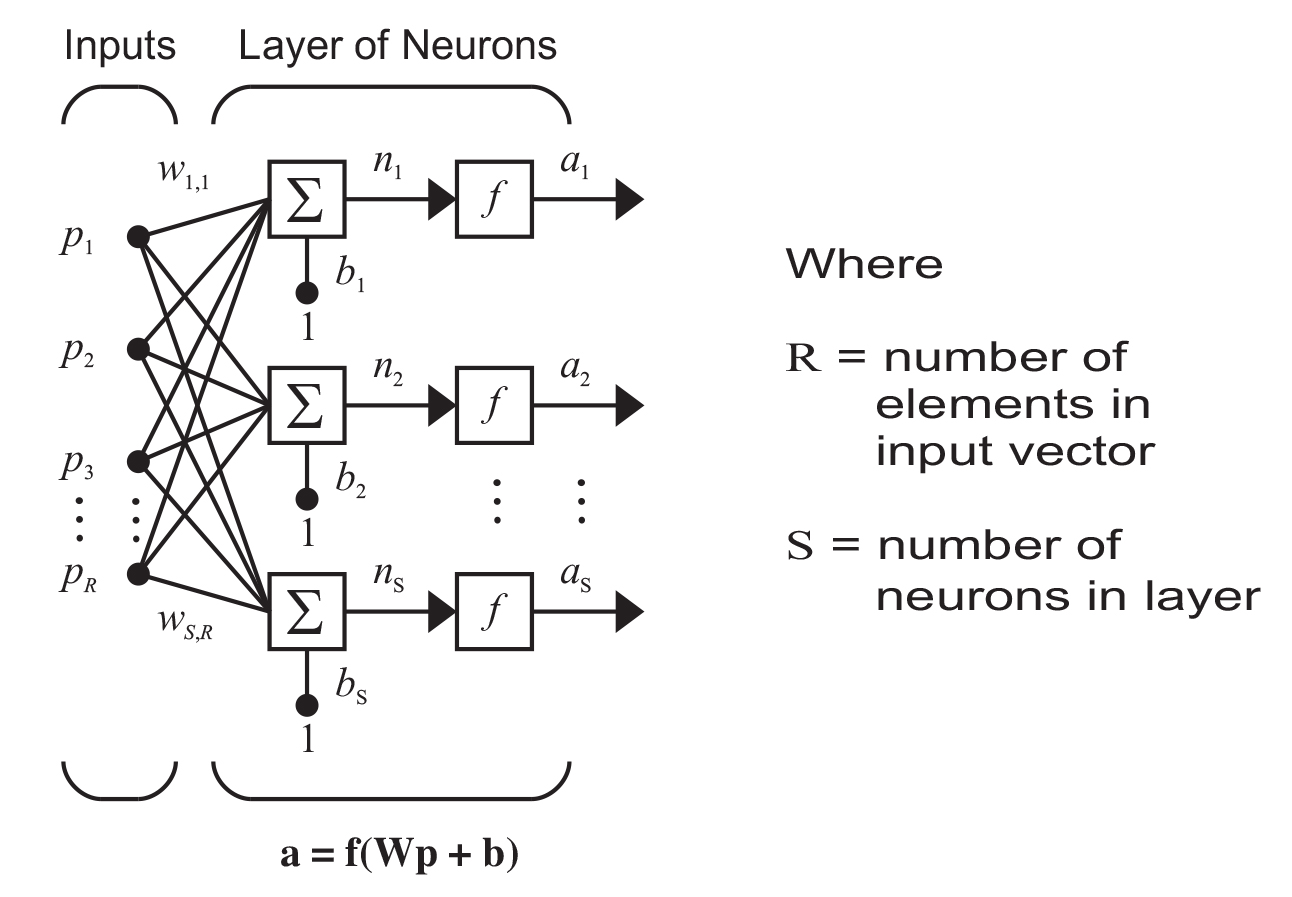}
    \caption{Diagram of the nodes, connections, weights, and biases in one layer of a neural network. \cite{neurons}}
    \label{fig:BasicNN}
\end{figure}
By stacking enough layers, where the output of one becomes the input of the next, any non-linear equation can be learned.

It was not until 2005, with the first application of General Purpose Graphics Processing Units~\cite{GPUs} (GPGPUs, commonly referred to as GPUs) did large or deep neural networks become practical.  Deep Neural Networks (DNNs) are loosely defined as networks with many hidden layers.  Even then, neural networks were still not well suited to processing image data, as evidenced by the results of the ImageNet classification challenge~\cite{ILSVRC15}.

Finally, in the year 2012, the use of a Convolutional Neural Network produced an error rate less than half of that of the next competing classifier (16\% vs  26\%)~\cite{ILSVRC15}. Following this, there was an explosion of interest in convolutional neural networks, and now they dominate the landscape of machine learning applications.

\TUsubsection{Possible Problems with NNs}
\label{sec:fit}
Neural Networks can suffer from several failure modes.  Most common are over- and under-fitting.  Over-fitting occurs when the network models the training data too well, including noise in the data, thus missing the overall pattern and suffering from poor generalization to unseen test data.  Over-fitting is typically caused by using too many parameters.  Under-fitting happens when the learned model misses major features, typically due to too few parameters, or insufficient training.
\begin{figure}
    \centering
    \includegraphics[width=0.75\linewidth]{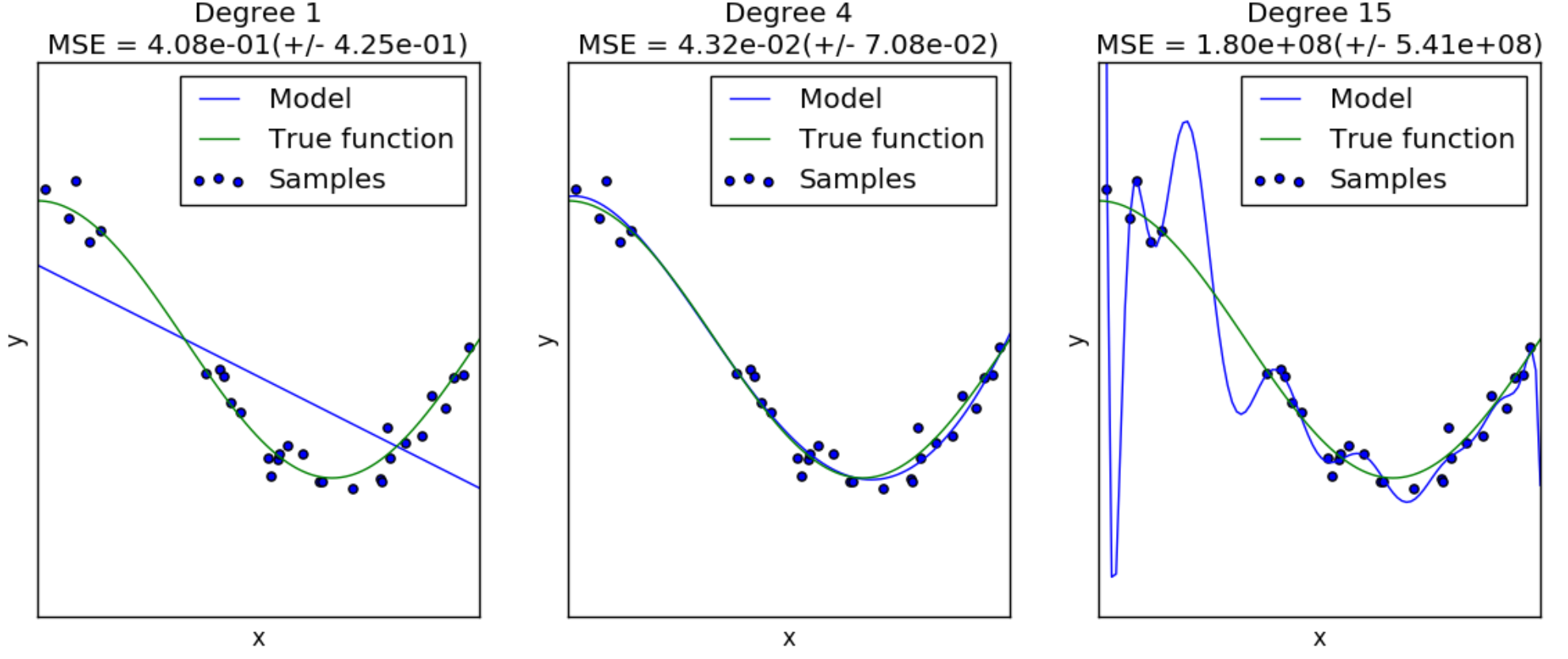}
    \caption{Demonstration of (left) under-fitting, (middle) correct fit, and (right) over-fitting \cite{scikit-learn}}
    \label{fig:fit}
\end{figure}

\TUsubsection{Loss Function}
A neural network is trained using a loss function on the outputs. NNs are frequently  trained on categorical data.
In these problems, a commonly used loss function is categorical cross-entropy performed on the output, computed using the softmax function of the final layer. The softmax function ``squashes'' a feature vector so that each item is in the range (0, 1) and the sum of the items is 1:
\begin{equation}
    \label{softmax}
    \sigma(z)_j = \frac{e^{z_j}}{\sum_{k=1}^{K}e^{z_k}},
\end{equation}
where \(j=1\ldots K\).

Categorical cross-entropy is a loss function which is minimized with the correct answer.  Categorical cross-entropy uses the equation
\begin{equation}
    \label{crossentropy}
    H(p,q)=-\sum_{j}p(j) \log q(j),
\end{equation}
where \(j=1\ldots K\), \(p(j)\) is the desired probability (1 for the correct category, 0 otherwise), and \(q(j)\) is the actual probability from softmax.  This loss uses the known information of the correct category to guide the gradients.  These gradients form the error term in equation \ref{eq:backprop}.

\TUsubsection{Convolutional Neural Networks}

The basis of a Convolutional Neural Network (or CNN) is the convolution operation. The convolution operation takes a small array called a kernel and passes it over the input.  Using equation~\eqref{eq:convolution} for each pixel \((i,j)\) in the input produces an output filtered by the kernel. Depending on how the borders are padded, the size of the output can be the same, or smaller than the input by up to half the kernel width:

\begin{equation}
    \label{eq:convolution}
    o(i,j) = \sum_{k,l = -w}^{+w}{K(k+w,l+w) * I(i+k,j+l)}.
\end{equation}

\begin{figure}[t]
    \centering
    \includegraphics[width=.75\linewidth]{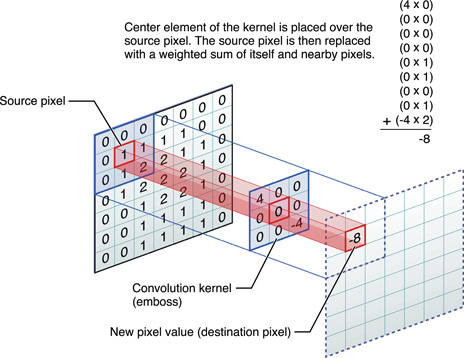}
    \caption{Graphical representation of 2D Convolution \cite{website:arm-community}}
    \label{fig:convolution}
\end{figure}

As employed in a neural network, convolution allows the use of a small number of weights across a large number of inputs.  Both training and prediction with a convolution are much faster than when using a different set of weights for each input pixel.  Where a fully connected layer for a small 128x128 image has 16384 weights per pixel for a total of more than 268 million weights, a convolutional operation of 5x5 has only 25 weights, regardless of image size.

The equations for a convolutional layer are very similar to a fully connected layer.  Where a fully connected layer has single values for its input and output, a convolutional layer has channels.  The convolution operation (\(\circledast\)) is performed in place of the weight multiplication:
\begin{equation}
    \label{eq:convlayer}
    C_i = \sum_{I\in Input}{(K_{iI}\circledast I)} + b_i.
\end{equation}
Just as each neuron in Equation \ref{eq:neuron} has one weight per input, each convolutional layer has one kernel per input channel.

\begin{figure}[ht]
    \centering
    \includegraphics[width=0.75\linewidth]{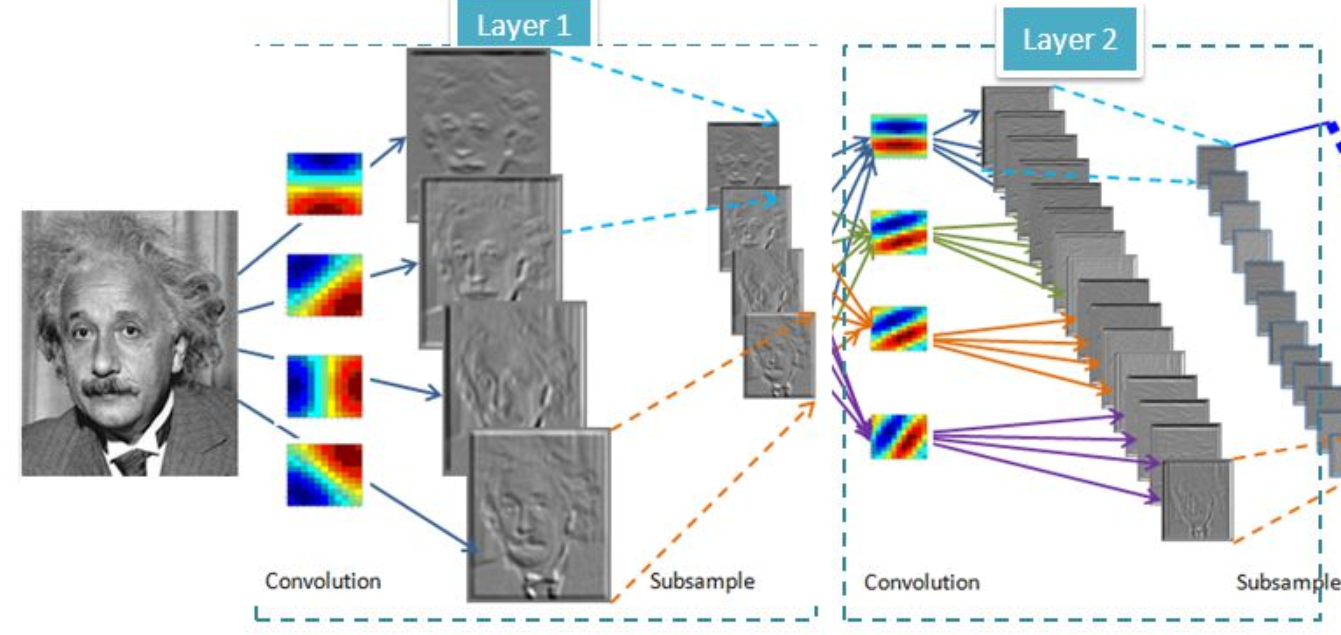}
    \caption{Diagram showing two layers of a Convolutional Neural Network \cite{einstein}}
    \label{fig:CNN}
\end{figure}

A convolutional layer has an output width and height that depend on parameters such as the padding and stride of the layer.  The most typical use case is padding the input with zeros with a stride of one to produce the same size output.  The depth of the output will be the same as the number of kernels (or filters) in the layer.  Since the input may have a depth greater than one, this is resolved by filtering each layer with each kernel, and the temporary responses are combined as if there were a pixel-wise fully connected layer.  In other words, the first output layer is the weighted sum of the first kernel applied to the input layers, plus a bias.

For example, the CIFAR-10 network described in Appendix~\ref{App:A} has an input of a 32x32x3 image: 32 by 32 pixels, and 3 colors.  The first layer of the network is defined a 3x3 convolution producing an output of 32x32x32 (Width x Height x Depth).  Connecting the input to each channel of the first layer are 3 convolutional operations, one for each input color.  Because the depth of the second layer is 32, there are 3*32=96 convolution operations. Because the convolution used for this layer is 3x3, there are 96*3*3=864 weights.  In addition, there are 32 biases, to make 896 total parameters.

\TUsubsection{Benefits of a CNN}

Common kernels such as the Laplacian and the Sobel kernel detect points and edges in an image, respectively. An important feature of these kernels is that they work just as well in all parts of the image, a property called \textbf{translation invariance}.  This property is extremely useful for convolutional neural networks.  A fully connected layer might learn to respond to a corner of the object, but only where the corner appears in the training set.  A convolutional layer can learn the corner and respond just as well at any location in the image, greatly reducing the training data necessary.

The second useful property of a CNN is called \textbf{local connectedness}.  This property means that the output response is affected only by nearby inputs.  Local connectedness is an intuitive consequence of the limited size of the kernel and is considered a desirable property because it matches how our brains see the world.

\TUsection{Motivation for Compression}
As deep neural networks become more effective and useful, the desire to use them in more applications grows.  The most common impediments against deep NNs are speed, memory usage, and power usage.  Several neural network compression algorithms have been developed using a variety of strategies to target each of these limitations.  

The limitations of speed, memory usage, and power usage are not independent.  Research into the causes of power usage~\cite{Power} show that DRAM memory access uses more than 6000 times as much power as an integer addition, and more than 150 times as much as floating point multiplication.  Thus, anything that reduces memory access will reduce power usage also.  In addition memory access is much slower than mathematical operations, so eliminating them can improve speed.  All else being equal, a compressed network uses fewer operations, which improves inference speed.

\TUsection{Compression Methods}

The area of network compression falls into three main categories.  Quantization is the process of representing each parameter of a network with a reduced bit rate, either by reducing the precision, employing a look-up table, or combining similar values.  When a network is quantized, extra computational time may be required to access a look up table or to undo an encoding to restore the original value.  Pruning methods excise nodes that have small effects, replacing them with null operations.  Pruning can produce sparse layers, which may take longer to perform than the original.  Lastly, retraining methods replace either the whole, or part, of the network with a freshly trained replacement.  Retraining incurs no run-time cost, but the upfront effort can be significant. 

Fortunately, these methods are not mutually exclusive.  As shown in \cite{Deep-Compression} and \cite{SqueezeNet}, a network that is pruned or retrained can then be quantized with little to no reduction in accuracy.  In addition, a mild form of retraining called fine tuning is used by most pruning methods to recover accuracy lost during the pruning process.  Some quantization methods can benefit from fine tuning as well.

\TUsubsection{Quantization}
\label{sec:Quantization}
Quantization methods include Huffman coding \cite{Van-Leeuwen}, reduced bit-depth networks \cite{Gong-et-al.}, and weight sharing \cite{Chen-et-al.}.  All of these reduce the numerical accuracy of the individual weights while ideally maintaining the same output results.  

\begin{figure}[ht]
    \centering
    \includegraphics{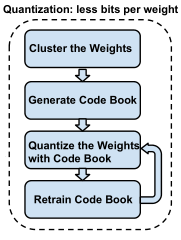}
    \caption{Example of Quantization steps from \cite{Deep-Compression}}
    \label{fig:Quantization}
\end{figure}
Huffman coding is a well known loss-less compression scheme that uses variable-length words to encode symbols.  By using smaller words for the most common symbols, the total space used is reduced.

Reduced bit-depth networks are created and trained from the beginning with fewer bits than normal.  Most common is FP-16, a 2 byte floating point format that is well supported by processor and GPU hardware.  Less common are format such as ms-fp8 and ms-fp9, which are 8- and 9-bit floating point formats from Microsoft \cite{Brainwave}.  No matter the physical format, all reduced bit-depth networks share the feature of trading numerical accuracy for space, and often for speed.  Fortunately, neural networks are resilient to such inaccuracies, and typically learn correct mappings.

Weight sharing is precisely what it sounds like, with multiple connections using the same weight.  For example, by limiting the number of weights to 256, each connection may be represented by 8-bits, instead of the typical 32.  In \cite{Deep-Compression}, convolutional layers can be represented with 256 shared weights, and fully connected layers with 32, with no loss of accuracy. In experiments, a network using 32-bit weights can be compressed to 6-bit weights with no loss in either Top-1 or Top-5 accuracy \cite{SqueezeNet}.

Some quantization schemes retain the ability to fine-tune the network by retraining with the reduced parameters.  Others work in a way that precludes using the typical gradient descent.  Weight sharing and reduced bit-depth methods are typically able to retrain, but more complex encoding schemes, such as Huffman coding, cannot be retrained.

\TUsubsection{Pruning}
\label{sec:Pruning}
Where quantization tries to reduce the space used for each connection, pruning methods work to reduce the number of connections.  Pruning methods typically work by analyzing the network for the weights with the least impact on the results and then removing them.  In addition to reducing network size, pruning can have benefits \cite{OBS-cite-7} such as reducing over-fitting \ref{sec:fit}.

The classic methods for pruning are Optimal Brain Damage (OBD) \cite{OBD} and Optimal Brain Surgeon (OBS) \cite{OBS}, both of which are computationally very expensive. OBD requires testing every input on every variable to calculate all the second derivatives, while OBS requires inverting an NxN Hessian matrix, where N is the number of weights, which for VGG-16 means N = 138 million.

Research in this area has focused on finding more computationally efficient methods, such as performing OBS in a layer-wise fashion \cite{L-W-OBS} or simply removing the values closest to zero, such as in \cite{Han-et-al.}.  Other methods are designed specifically to prune convolutional layers.  One removes the filters with the smallest \(\ell_1\)-norm \cite{Pruning-Filters}.  Another method is found in \cite{ThiNet} where a greedy algorithm minimizes the error of removed filters.

\TUsubsection{Retraining}
\label{sec:Retraining}
Retraining methods initialize the network from scratch with fewer weights and/or a different architecture.  Strictly speaking, any new architecture is a form of retraining, such as \cite{SqueezeNet}, which uses different layer types and bypass layers. In the context of network compression, though, it refers to methods that reuse some parameters, whole layers, other information from an original network.

Knowledge Distillation \cite{Knowledge-Distillation} is a method for training a network based on the output of another network. Its descendant, FitNets \cite{FitNets}, uses the output from a single intermediate layer in addition to the network's final output.  The ThiNet method \cite{ThiNet} divides a network into sections and uses the output of each section to retrain the pruned layers earlier in the section.

\TUsection{Why another method?}
The current state-of-the-art neural network compression methods include {\tt Learning Both Weights and Connections}~\cite{LBWC}, {\tt ThiNet}~\cite{ThiNet}, and {\tt FitNets}~\cite{FitNets}.  Each of these has its unique strengths and weaknesses.

\TUsubsection{Learning Both Weights and Connections}
\label{sec:LBWC}
Learning Both Weights and Connections (LWBC) is described in~\cite{LBWC}, \cite{Deep-Compression}, and~\cite{Han-et-al.}.  It works by pruning all weights below a certain threshold, followed by fine tuning the weights of the now sparse network.  This is repeated for five iterations.  

The LWBC method has several advantages, primarily its simplicity and low computational burden, as it requires retraining only once per iteration.  On the other hand, it produces a sparse network, which is a large disadvantage.  Such networks require specialized hardware to function efficiently, and are incompatible with normal neural network software.

\begin{figure}[ht]
    \centering
    \includegraphics{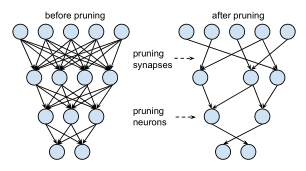}
    \caption{Illustration of the LBWC process. \cite{LBWC}}
    \label{fig:LBWC}
\end{figure}

Another disadvantage is that the network architecture is static.  Architecture can have a very large effect on accuracy~\cite{SqueezeNet}, and the inability to alter it restricts the possibilities for improvements in this area.

\TUsubsection{ThiNet}
\label{sec:ThiNet}
The ThiNet method \cite{ThiNet} resolves some of the problems with LBWC.  Beginning with the first layer, a value is calculated for each channel.   After removing the least valuable channels to reach a desired size or accuracy, the layer being pruned and the layer after it are retrained.  Then the process repeats for the next pair of layers.  The dense network produced is more effective than LBWC because it can be used with normal software and hardware.

\begin{figure}[ht]
    \centering
    \includegraphics{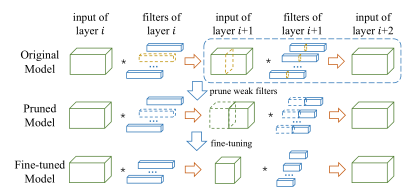}
    \caption{Illustration of the ThiNet process. \cite{ThiNet}}
    \label{fig:ThiNet}
\end{figure}

As with LBWC, the architecture of ThiNet is static.  However, operating in a layer-wise fashion allows more flexibility to explore different network shapes.  For example, layers with an out-sized importance on the overall accuracy can be slightly compressed, while less important ones can be strongly compressed.  

\TUsubsection{FitNets}
\label{sec:FitNets}
The FitNet method \cite{FitNets} is a retraining based method.  Like Knowledge Distillation \cite{Knowledge-Distillation}, it uses the information from a teacher network to assist the training.  FitNets tries to solve the vanishing gradient problem and speed up training by including an additional``hint'' layer.  This layer is a point of correspondence between the new and the teacher networks that is included in the loss function (described in \ref{sec:Loss}).

\begin{figure}[ht]
    \centering
    \includegraphics[width=0.75\linewidth]{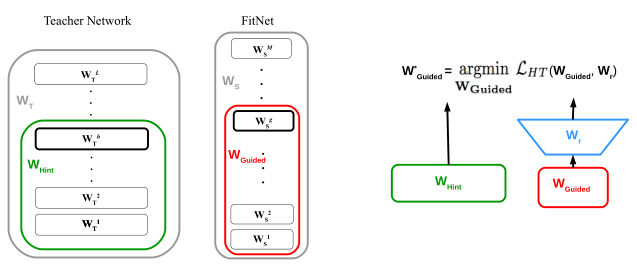}
    \caption{Illustration of the FitNets process. \cite{FitNets}}
    \label{fig:FitNets}
\end{figure}

While this method does have the advantage of being able to alter the architecture, its possibilities are restricted by requiring the hint layer to match a layer in the original architecture.  In addition, exploration of the space requires retraining the entire architecture, which takes a significant amount of time.

Another possible disadvantage of FitNets is the effect of the ever-increasing depth of networks.  The hint layer is supposed to reduce the problems with deep networks, but the deepest one tested was 19-layers, which is shallow by the standards of ResNet \cite{ResNet}, which has been tested out to more than 1000 layers, though ResNet-50 is the most common implementation.

\TUsection{Desired properties}
\label{sec:CramNet}
Any new network compression algorithm should address the weaknesses of previous methods.  It should produce a dense network, to ensure it improves speed and size.  It should be able to reshape the network to take advantages of the latest in architectures.  Most importantly, it should compress the network as well or better than existing methods without reducing the network accuracy.

This chapter showed the basics of neural networks and convolutional neural networks, and explained how convolutional neural networks can be better for image data.  Then it discussed the types of neural network compression, as well as several existing compression methods.  The next chapter will explain the CramNet method and how it is different from the previous methods.

\TUchapter{CramNet}

This thesis explores a new neural network compression method called CramNet.  The objective of CramNet is to improve the architectural flexibility of FitNets, while combining it with the easy exploration and dense network of ThiNet, all without sacrificing acccuracy.  The result should be smaller, more flexible networks suitable for mobile and time-constrained applications.

\TUsection{Compression Procedure}

The optimization technique of CramNet is designed to be performed on a per-layer basis by breaking it down into sub-problems.  Having a set of sub-problems allows the compression to halt when a desired compression, or a certain loss of accuracy, is reached.  Each sub-problem is solved in ``reverse'' order, from the output to the input.

\begin{figure}[ht]
    \centering
    \includegraphics[width=1\linewidth]{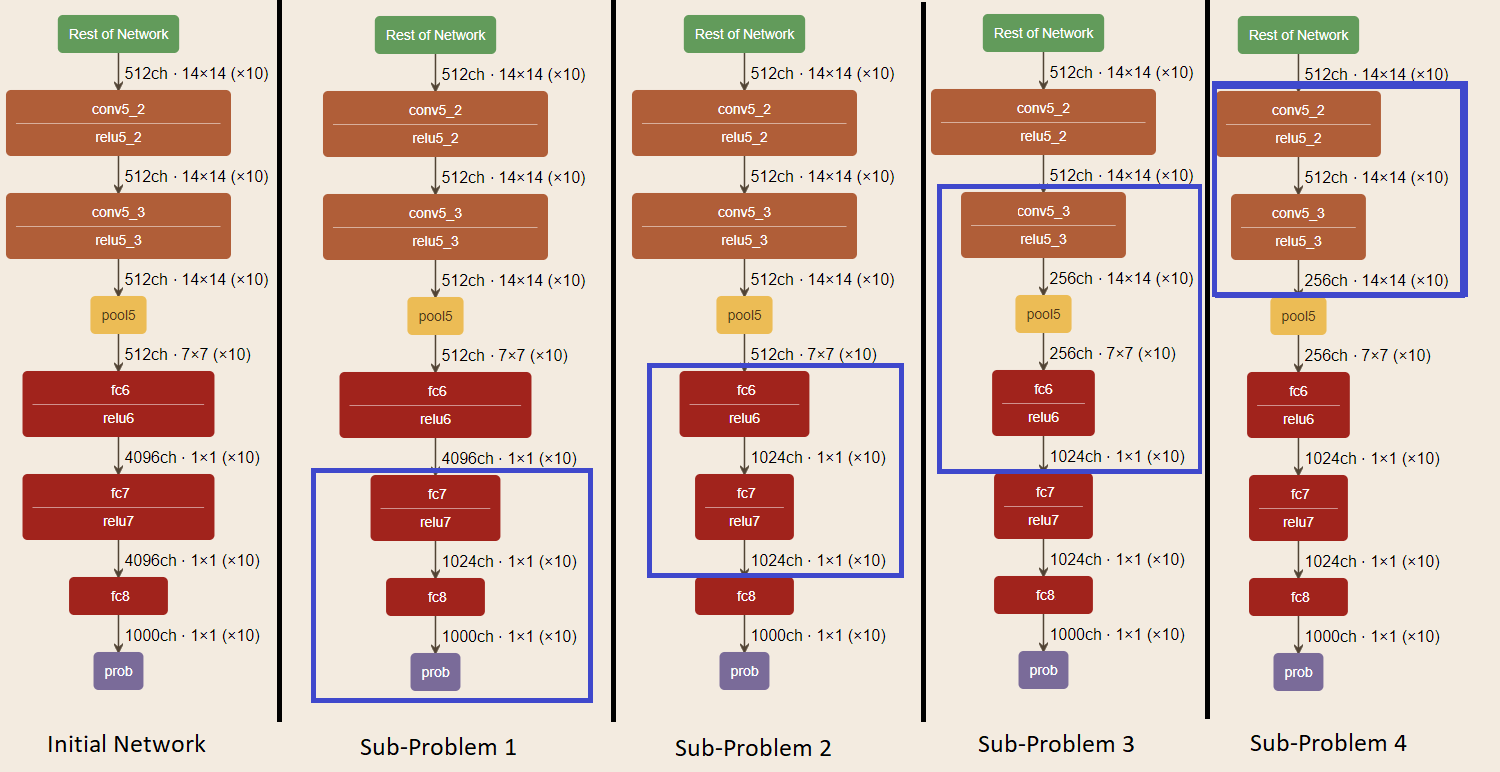}
    \caption{A similar architecture, with sub-problems outlined in blue. Initial network is on the left, progressing to the right.}
    \label{fig:Procedure}
\end{figure}

\TUsubsection{Sub-Problem Definition}

The network is divided into sections, typically containing two sets of parameters.  For example, in Figure \ref{fig:Procedure}, sub-problem 1 and 2 each consist of three fully connected layers, between each of which is a set of parameters.  Sub-problem 3 is one convolutional layer, one max pooling layer, and two fully connected layers, because the max pooling layer includes no parameters.  For each of these sub-problems, the first layer provides the input, sized to match the original network.  The second layer is being re-sized or replaced, which alters the number of parameters connecting it to the layers before and after it.  The third layer is sized to match the rest of the compressed network.

Again referring to Figure \ref{fig:Procedure}, sub-problem 1 has an input of 4096 nodes, an output of 1000 nodes, and re-sizes layer fc8 from 4096 to 1024 nodes. Sub-problem 2 has an input of 25,088 nodes, and outputs to the new fc8 layer with 1024, while re-sizing layer fc7 from 4096 to 1024 nodes. Sub-problem 3, has an input of 512 channels, and outputs to the new fc7 with 1024 nodes.  To re-size the convolutional layer conv5\_3, the number of channels is reduced from 512 to 256.

\TUsubsection{Training Different Architectures}

FitNets (in \ref{sec:FitNets}) is capable of training an architecture that is different from the original teacher network. However, any to that architecture requires re-training the entire network from scratch.  In contrast, the sub-problem arrangement of CramNet means that each change means re-training a single sub-problem.  

An example of this would be replacing one or more of the convolutional layers with a residual block as described in \cite{ResNet}.  Residual blocks with a bypass show an improved resiliency to the vanishing gradient problem and use fewer channels to achieve the same or better accuracy.  An obvious test would be to explore how many channels are optimal.  Where FitNets would need to optimize the depth of all its layers at once, CramNet can optimize one sub-problem at a time.

\begin{figure}[ht]
    \centering
    \includegraphics[width=0.5\linewidth]{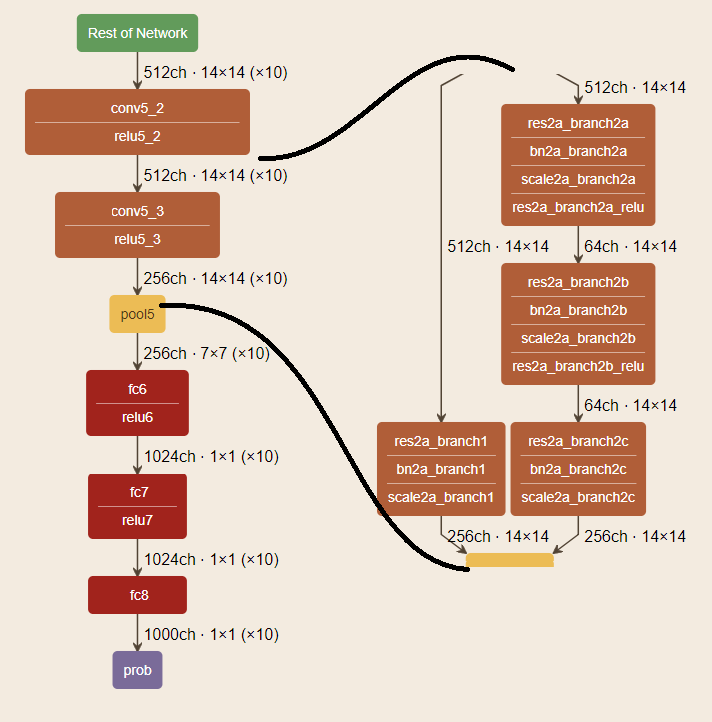}
    \caption{Sub-problem 4, showing the replacement of a convolutional layer with a residual block.}
    \label{fig:Procedure_Res}
\end{figure}

Sub-Problem 4 can be modified to replace the convolutional layer conv5\_3 with a residual block with no modification, as shown in  Figure \ref{fig:Procedure_Res}.  More complicated architectures such as bypass or branches that break sub-problem boundaries require more care.  When training the layer receiving the bypass (which may not yet exist, as it is earlier in the network), use a random noise sample as the input.  Then, when training the layer that produces bypass, the parameters from the receiving layer also need to be retrained.

\TUsection{Training}

With the sub-problem prepared, training goes much like any other network.  First, the altered origin network predicts data, producing its original output and an additional output that serves as input to the sub-problem.  The CramNet sub-problem is then trained using the additional output as its input, with the raw output and true labels as the target.  Sub-problems cannot be trained in parallel, as each depends on those that came before. 

\begin{figure}[ht]
    \centering
    \includegraphics[width=0.5\linewidth]{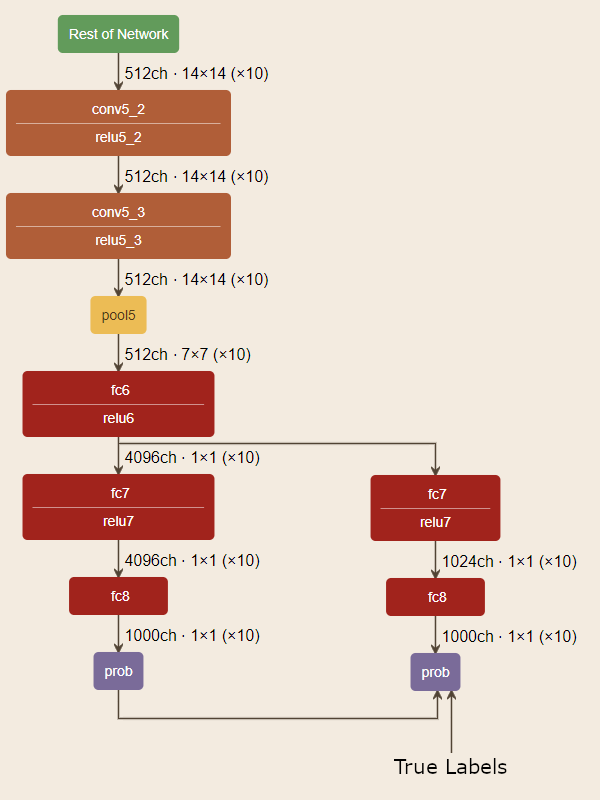}
    \caption{The training of Sub-Problem 1, showing the inputs and outputs.}
    \label{fig:Procedure_Training}
\end{figure}

Convergence happens much faster than for a full network, since only a few layers of the CramNet are trained at any one time.  As with most compression methods, a final fine-tuning should be performed to tighten up the results.

\TUsubsection{Optimizer}
All experiments in this paper use the RMSProp optimizer~\cite{RMS}.  Early tests show that most other optimizers available either fail to converge, or fail to alter the weights of the network at all.  These failures are likely due to the vanishing gradient problem, which is when the gradient becomes smaller the farther the parameter is from the outputs used in the loss function.

RMSProp helps mitigate this effect by dividing ``the learning rate for a weight by a running average of the magnitudes of recent gradients for that weight.'' \cite{RMS}  This is the batch equivalent of simply using the sign of the gradient, rather than its magnitude.

\TUsubsection{Order of Optimization}
A question that came up early in our research was whether optimization should begin from the output layer towards the input, or from the input to the output.  Intuitively, starting at the output is better for most networks, as the final fully connected layers contain most of the nodes.  

An experiment was performed using training a slightly compressed network from both directions.  The test was conclusive.  Training the network from the output end succeeded with accuracy comparable to the original network.  On the other hand, training from the input end produced steadily degrading results.  With each layer, the accuracy reduced, until the loss diverged to infinity and accuracy, to zero!

\TUsubsection{Loss Function}
\label{sec:Loss}
Where a normal network can only train on the true labels, when CramNet is compressing it can use the output from the teacher network.  To take advantage of that extra information, some modifications to the standard equations must be made.

To use the full array of category probabilities when training, a different loss function is needed.  The appropriate loss for this case is the mean squared error, for which the equation is
\begin{equation}
    \label{mse}
    E(p,q) = \frac{1}{K} \sum_{j} (p(j)-q(j))^2 | j=1...K
\end{equation}
where \(p(j)\) is the pre-softmax value for category j in the teacher network, and \(q(j)\) is the pre-softmax value for category j in the network being trained.  However, this is not the optimal loss.

To reach the best overall accuracy, the training must be guided not only by the teacher network, but also by the true data.  Thus, as described in \cite{Knowledge-Distillation} and \cite{FitNets}, the optimal loss is 
\begin{equation}
    \label{eq:CramNet-Loss}
    L(p,q,l) = E(p,q) + H(l,\sigma(q))
\end{equation}
where \(p\) is the output of the teacher network before softmax, \(q\) is the output of the student network before softmax, and \(l\) contains the true labels.  \(E(p,q)\) is \eqref{mse}, and \(H(p,q)\) is \eqref{crossentropy}, and \(\sigma(q)\) is \eqref{softmax}.

The loss presented in \ref{eq:CramNet-Loss} is more effective than either \eqref{mse} or \eqref{crossentropy} individually because it uses the information from the teacher network to help avoid local-minimums and still pushes towards the correct answer.  

This chapter explained the CramNet method.  Beginning with the overall procedure, it showed how to define sub-problems, and how to alter them for different architectures.  Then it explained the technical details of how to set up the optimization problem, from the optimizer method to use, the loss function and even which order to solve the sub-problems in.

\TUchapter{Experiments}

This chapter is dedicated to describing the experiments performed to validate the CramNet model and the corresponding results.  First, the metrics used to compare CramNet with itself and other methods are described.  CramNet is validated on two datasets: CIFAR-10 and ImageNet.  For each testset, the CramNet tests are described, then the results are compared to other methods, such as ThiNet and FitNets.

All experiments were performed using a single NVIDIA 1080 Ti GPU with 11GB of memory.  The networks were modeled and trained using the Keras Python deep learning library \cite{Keras}.

\TUsection{Evaluation Metrics}
\label{sec:Metrics}
The comparison of CramNet to other compression methods is a difficult task.  Every method may start with a different network because there is no standard network to use as a benchmark.  For example, the network used as the teacher in \cite{FitNets} has approximately 9 million parameters, whereas the network used for CramNet experiments, see (\ref{sec:Architecture}), has only 1.25 million.  In addition, the accuracy of the teacher networks differ by almost 10\%, 90.18\% for \cite{FitNets} and 80.64\% for the network used by CramNet.

In an attempt to normalize the performance of different algorithms across the variety of networks, three metrics were chosen. 

\TUsubsection{Compression Metric}
 The first metric is the \% of original parameters in the compressed network.  This is the ratio \(\frac{\# params_{new}}{\# params_{old}}*100\%\).  Normalizing the value and removing the dependence on initial network size clarifies the effect of the algorithms in isolation.

\TUsubsection{Accuracy Metric}
To compare the effect of the compression algorithms upon the accuracy, a similar procedure is performed.  The accuracy of a network is typically expressed as the percentage of the test set correctly classified.  Some papers use the error percentage, which is simple to convert to the accuracy, by subtracting it from 100\%.

The initial accuracy, \(a_{100}\), is the accuracy of the initial network on the dataset.  The compressed accuracy, \(a_c\) is the accuracy of a compressed network.  The most relevant value is the change in accuracy caused by the compression, defined as \(\Delta a = a_c - a_{100}\).  

This metric is not normalized by \(a_{100}\) for two reasons.  First, if \(a_{100}\) were very small, the network chould show an improvement of over 100\%, which is obviously impossible.  Secondly, no papers were found that normalized the change in accuracy by the initial accuracy.  Since there is no good reason for normalizing the metric, it was kept un-normalized to maintain easy comparison with other methods.

\TUsubsection{Performance Metric}

The last metric is the effect on performance.  Though true performance is heavily dependent on the hardware and software architecture, a decent proxy is the number of Floating Point Operations (FLOPs).  Since different initial networks have different starting values, the metric is normalized by the initial value, much like the Compression Metric.  The ratio is \(\frac{\# FLOPs_{new}}{\# FLOPs_{old}}*100\%\).  This percentage can be used to compare different algorithms by removing dependency on the complexity of the original network.

\TUsection{Datasets}

\TUsubsection{CIFAR-10}

The CIFAR-10 dataset \cite{CIFAR-10} contains 10 classes of 32x32 pixel RGB images, with 50 thousand training, and 10 thousand test images.  With a small number of clearly separated classes, and a relatively small number of images, networks can be trained and tested quickly.

\begin{figure}[]
    \centering
    \includegraphics[width=.5\linewidth]{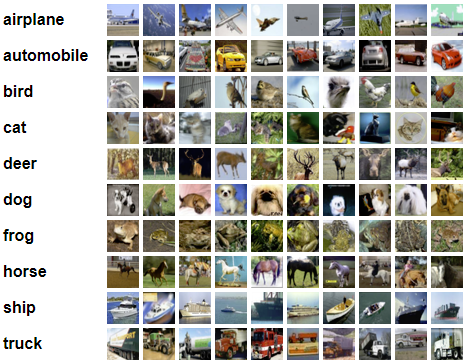}
    \caption{The categories of the CIFAR-10 dataset, with 10 random images from each category~\cite{CIFAR-10}.}
    \label{fig:CIFAR-10}
\end{figure}

The networks are evaluated on the 10,000-item test set with a standard categorical accuracy metric. The use of a dedicated test set makes CramNet's results using CIFAR-10 directly comparable with most other networks that use the same dataset.

The CIFAR-10 dataset is one of the chosen datasets for several reasons.  It is well-constructed and easy to use.  The dataset is packaged neatly, with well-defined training and test sets.  Therefore, any variations in results are due to the influence of the networks and training.  Because of its good qualities, it is used in hundreds, if not thousands of papers, and thus there are many data points for comparison.

\TUsubsection{ImageNet}
The ILSVRC 2012 dataset, commonly known as ImageNet, is the gold standard for image classification datasets.  The ILSVRC 2012 competition was the first to conclusively demonstrate the power of convolutional neural networks to a large population.  Since then, nearly every new network architecture or training method has been tested against this dataset.

With 1000 categories and more than 1.2 million hand-labeled images, it is one of the most extensive datasets in common use.  The dataset is separated into training, validation, and test sets.  However, unlike CIFAR-10, the images are not all normalized.  The image sizes are not uniform.  Some networks crop the center of the images, others choose random translations, and still others re-size the images to the input size.

\begin{figure}[ht]
    \centering
    \includegraphics[width=.75\linewidth]{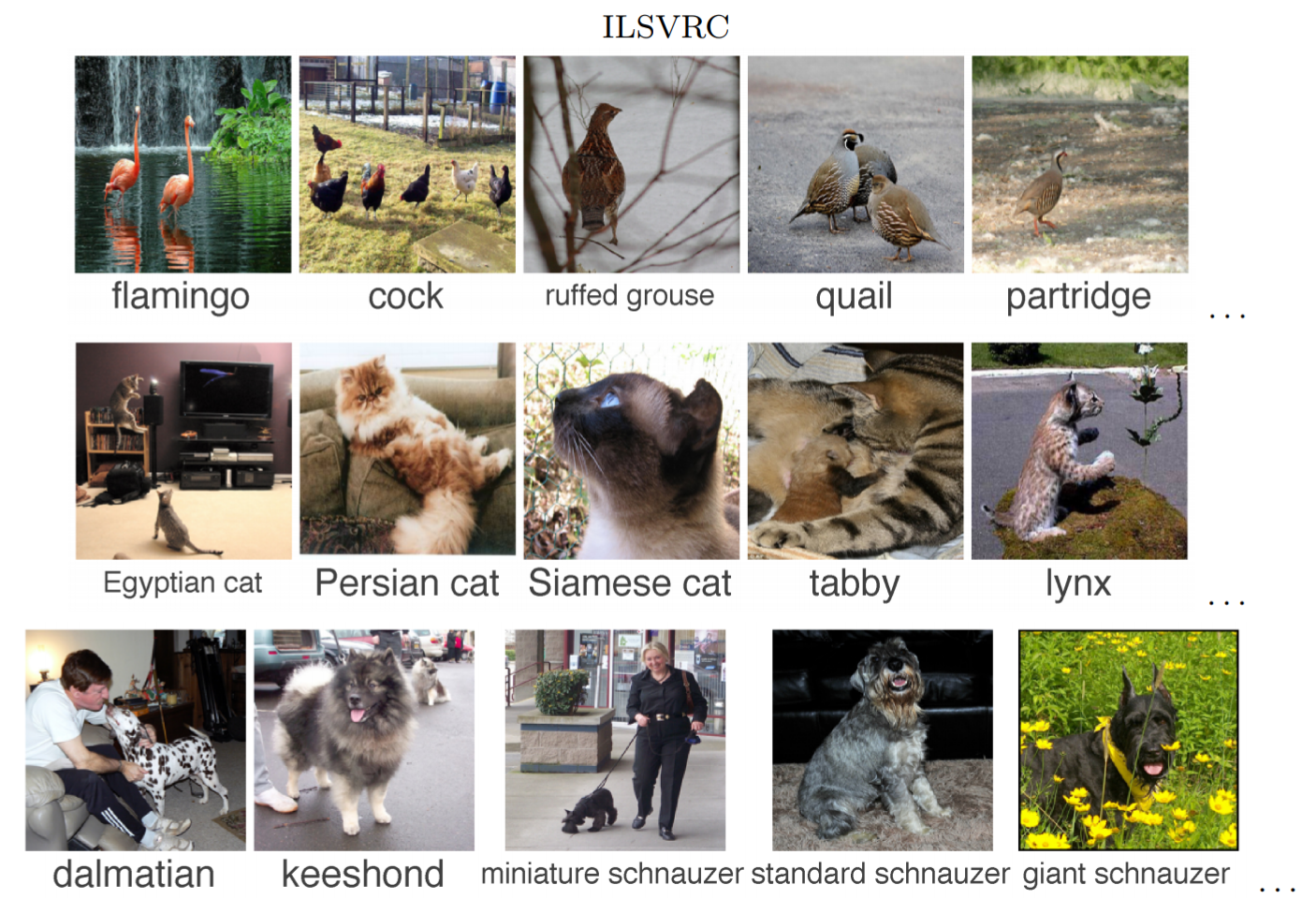}
    \caption{Several categories of the ImageNet dataset, showing the fine-grained class divisions. \cite{ILSVRC15}}
    \label{fig:ImageNet}
\end{figure}

\TUsection{Architectures Used for Experimentation}

\TUsubsection{CIFAR-10}
\label{sec:Architecture}
A network with proven effectiveness on the CIFAR-10 dataset was chosen to be compressed with CramNet.  As the basic CIFAR-10 example in the Keras neural network library \cite{Keras}, the architecture, shown in \hyperref[App:A]{Appendix~\ref*{App:A}} and Figure \ref{fig:DefaultArch-10}, is well tested and documented. It is simple, consisting of convolution layers interspersed with max pooling, followed by a fully connected layer.  All told, the architecture contains just over 1.25 million trainable parameters.

\begin{figure}[]
    \centering
    \includegraphics[width=1\linewidth]{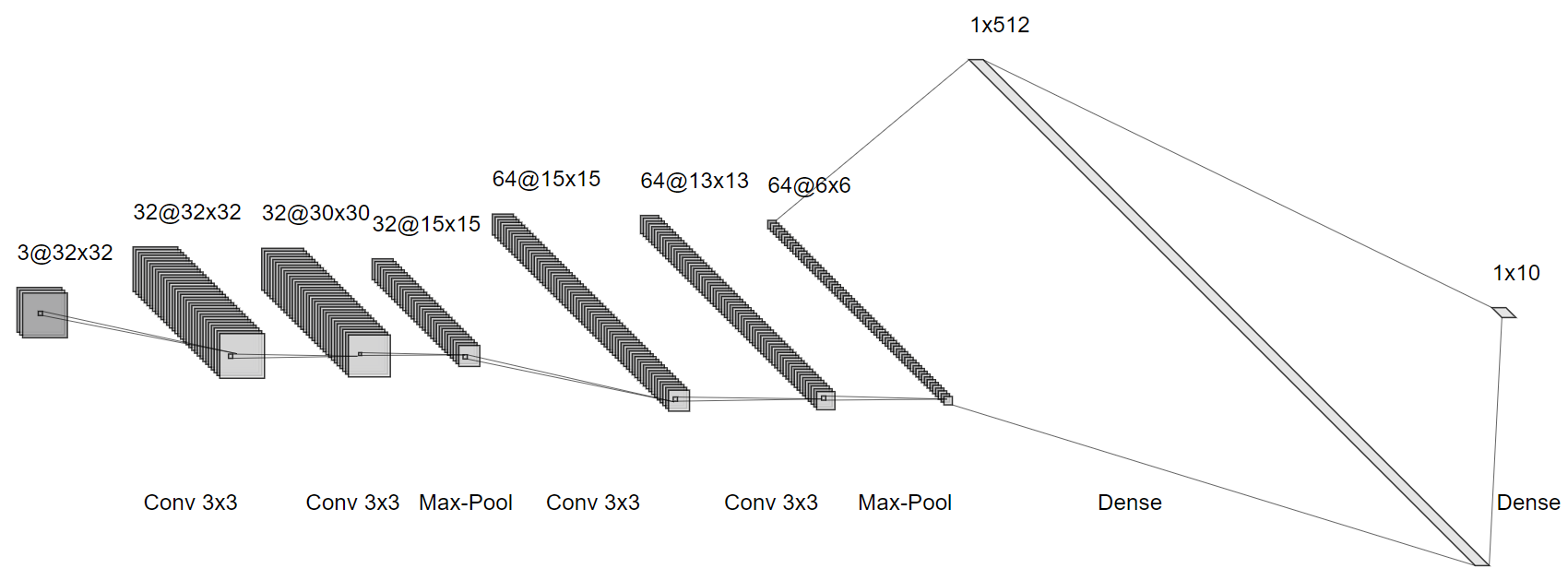}
    \caption{The Keras example architecture for the CIFAR-10 dataset.}
    \label{fig:DefaultArch-10}
\end{figure}

Trained using the RMSprop optimizer for 500 epochs, it achieves slightly more than 80\% accuracy.  This is the starting point against which CramNet was tested.  To avoid confusion with alternate factors, the architectures tested are merely ``thinner'' versions, with the same number of layers, and fewer fully connected nodes or filters per layer.

\TUsubsection{ImageNet}

There are several well-known and freely available networks for the ImageNet dataset.  The most common of these is the VGG-16~\cite{VGG}, which has 13 convolutional layers and 3 fully connected layers with more than 138 million parameters between them.  Other networks include the ResNet family and AlexNet.  The VGG-16 network was chosen primarily for its ubiquity.  Rare is the network compression paper that does not test using the VGG-16 network.

\begin{figure}[ht]
    \centering
    \includegraphics[width=1\linewidth]{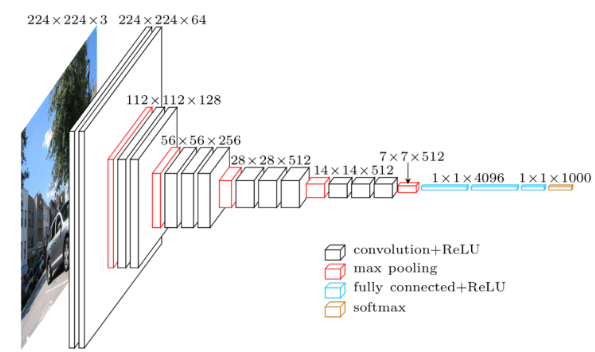}
    \caption{The VGG-16 architecture for the ImageNet dataset \cite{VGG-fig}.}
    \label{fig:VGG-16}
\end{figure}

Described in detail in Appendix \ref{App:B} and shown in Figure \ref{fig:VGG-16}, the VGG-16 network is both an architecture and a pre-trained set of parameters.  By starting with the exact same network and parameters, every method can be directly compared.  

Unfortunately, not all methods use the same metrics.  Several of the methods found during research provide only the percentage of FLOPs (FLOating Point Operations) that the method removes, rather than the parameter count.

The baseline accuracy of the network is 68.34\%, using over 138 Million parameters.  This is the starting point against which CramNet, and the other methods, are judged.  For ease of comparison, the normalized accuracy metric will continue to be used.

\TUsection{CramNet Results}

\TUsubsection{CIFAR-10 Results}

\begin{table}[ht]
\centering
\begin{tabular}{|c|c|c|c|c|}
    \hline
    Test \# & \# Parameters & \% of Parameters & \% of FLOPs & Accuracy Difference (\(\pm\%\))  \\ \hline
    \(\emptyset\) & 1,250,858 & 100 & 100 & +0 (Baseline) \\ \hline 
    1 & 128,314 & 10.26 & 25.17 & +1.02 \\ \hline 
    2 & 91,098 & 7.28 & 24.98 & +0.21 \\ \hline 
    3 & 72,490 & 5.80 & 24.93 & -0.6 \\ \hline 
    4 & 53,882 & 4.31 & 24.83 & -1.35 \\ \hline 
    5 & 35,274 & 2.82 & 24.83 & -3.38 \\ \hline 
    5b & 13,698 & 1.10 & 6.75 & -14.97 \\ \hline 
    1b & 60,658 & 4.85 & 6.94 & -5.52 \\ \hline 
    1c & 68,394 & 5.47 & 21.17 & -1.28 \\ \hline 
\end{tabular}
\label{table:CIFAR-10}
\caption{CramNet Results for the CIFAR-10 dataset}
\end{table}

For the CIFAR-10 dataset, there were 8 experiments.  Ranging from 1 to 11\% compression of the original network size, they demonstrate the change in accuracy with network size and architecture.

\begin{description}
\item[Experiment 0] Original network; Four convolutional layers, two fully connected layers with 512 intermediate nodes and 10 outputs
\item[Experiments 1] Half-depth convolutional layers, 96 intermediate nodes
\item[Experiments 1b] Quarter-depth convolutional layers, 96 intermediate nodes
\item[Experiments 1b] Three half-depth convolutional layers, one quarter-depth convolution layer, 96 intermediate nodes
\item[Experiments 2] Half-depth convolutional layers, 64 intermediate nodes
\item[Experiments 3] Half-depth convolutional layers, 48 intermediate nodes
\item[Experiments 4] Half-depth convolutional layers, 32 intermediate nodes
\item[Experiments 5] Half-depth convolutional layers, 16 intermediate nodes
\item[Experiments 5b] Quarter-depth convolutional layers, 16 intermediate nodes
\end{description}

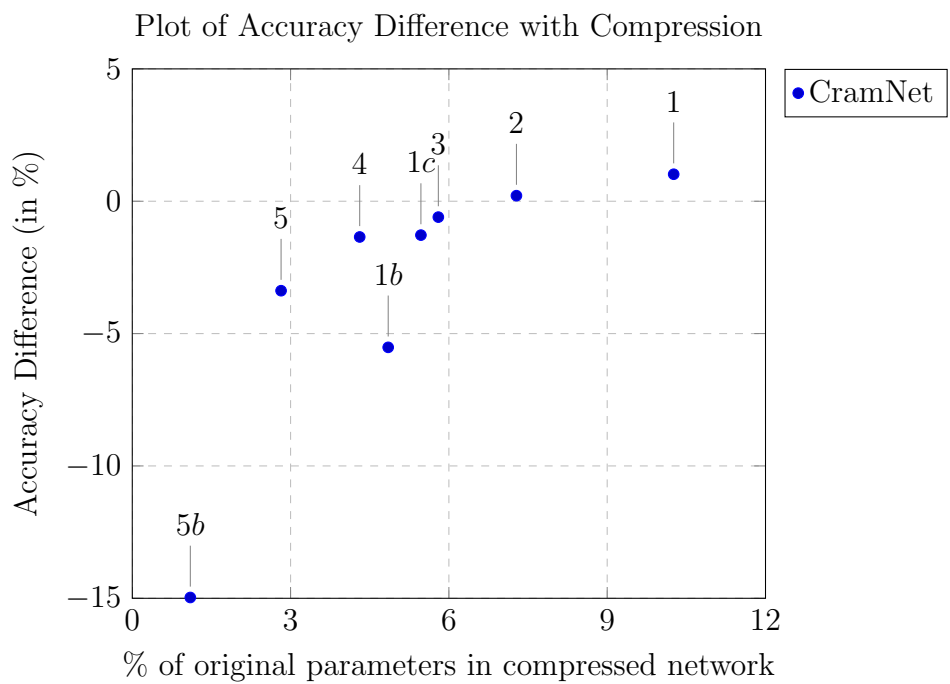
\begin{figure}[ht]
\centering
\begin{tikzpicture}
\begin{axis}[
    title={Plot of Accuracy Difference with Compression},
    xlabel={\% of original parameters in compressed network},
    ylabel={Accuracy Difference (in \%)},
    xmin=0, xmax=12,
    ymin=-15, ymax=5,
    xtick={0,3,6,9,12},
    ytick={-15,-10,-5,0,5},
    legend pos=outer north east,
    xmajorgrids=true,
    ymajorgrids=true,
    grid style=dashed,
    ]
\addplot+[
    only marks,
    ]
    coordinates {
    (10.26, 1.02)
    (7.28, 0.21)
    (5.80, -0.6)
    (5.47, -1.28)
    (4.85, -5.52)
    (4.31, -1.35)
    (2.82, -3.38)
    (1.10, -14.97)
    };
\addlegendentry{CramNet}
\addplot[mark=none] coordinates {(10.26, 1.02)} node[pin=90:{$1$}]{} ;
\addplot[mark=none] coordinates {(7.28, 0.21)} node[pin=90:{$2$}]{} ;
\addplot[mark=none] coordinates {(5.80, -0.6)} node[pin=90:{$3$}]{} ;
\addplot[mark=none] coordinates {(5.47, -1.28)} node[pin=90:{$1c$}]{} ;
\addplot[mark=none] coordinates {(4.85, -5.52)} node[pin=90:{$1b$}]{} ;
\addplot[mark=none] coordinates {(4.31, -1.35)} node[pin=90:{$4$}]{} ;
\addplot[mark=none] coordinates {(2.82, -3.38)} node[pin=90:{$5$}]{} ;
\addplot[mark=none] coordinates {(1.10, -14.97)} node[pin=90:{$5b$}]{} ;
\end{axis}
\end{tikzpicture}
\caption{CramNet Results for the CIFAR-10 dataset}
\label{fig:CIFAR-10_Results_CN}
\end{figure}

As Figure \ref{fig:CIFAR-10_Results_CN} shows, tests 1 through 5 (see \ref{A:Test 1} through \ref{A:Test 5} for specifics) produce a relatively smooth curve.  Interestingly, it seems that the network can be compressed to as much as 6-7\% of its original size before it begins to significantly lose accuracy.  The other tests demonstrate that it is not primarily the number of parameters, but is the architecture that impacts the accuracy of the network.  The wrong architecture, seen in test 1b and 5b, can triple the error over the basic tests.  

Test 5b (\ref{A:Test 5b}) uses quarter-depth conv2d with the same fc as Test 5, and 1b is the same for Test 1.  Using these end-points the effect on the other tests can be extrapolated.  

Test 1c is part of a set with 1 and 1b.  Together, they show the possible benefits of different arrangements of parameters. All three have a 96-node fc layer, but Test 1c (\ref{A:Test 1c}) has fewer quarter-depth conv2d layers than 1b (\ref{A:Test 1b}).  Almost as many parameters were removed, but the accuracy loss was more than halved.  These tests suggest that when exploring the parameter space, an iterative approach can be use, with experimentation stopping at the desired loss of accuracy.

\begin{figure}[ht]
\centering
\begin{tikzpicture}
\begin{axis}[
    title={Plot of Accuracy Difference with Compression},
    xlabel={\% of original FLOPs in compressed network},
    ylabel={Accuracy Difference (in \%)},
    xmin=0, xmax=30,
    ymin=-15, ymax=5,
    xtick={0,5,10,15,20,25,30},
    ytick={-15,-10,-5,0,5},
    legend pos=outer north east,
    xmajorgrids=true,
    ymajorgrids=true,
    grid style=dashed,
    ]
\addplot+[
    only marks,
    ]
    coordinates {
    (25.17, 1.02)
    (24.98, 0.21)
    (24.93, -0.6)
    (21.17, -1.28)
    (6.94, -5.52)
    (24.83, -1.35)
    (24.83, -3.38)
    (6.75, -14.97)
    };
\addlegendentry{CramNet}
\addplot[mark=none] coordinates {(25.17, 1.02)} node[pin=70:{$1$}]{} ;
\addplot[mark=none] coordinates {(24.98, 0.21)} node[pin=70:{$2$}]{} ;
\addplot[mark=none] coordinates {(24.93, -0.6)} node[pin=70:{$3$}]{} ;
\addplot[mark=none] coordinates {(21.17, -1.28)} node[pin=70:{$1c$}]{} ;
\addplot[mark=none] coordinates {(6.94, -5.52)} node[pin=70:{$1b$}]{} ;
\addplot[mark=none] coordinates {(24.83, -1.35)} node[pin=70:{$4$}]{} ;
\addplot[mark=none] coordinates {(24.83, -3.38)} node[pin=70:{$5$}]{} ;
\addplot[mark=none] coordinates {(6.75, -14.97)} node[pin=70:{$5b$}]{} ;
\end{axis}
\end{tikzpicture}
\caption{CramNet Results for the CIFAR-10 dataset using \% FLOPs metric}
\label{fig:CIFAR-10_Results_FLOP}
\end{figure}
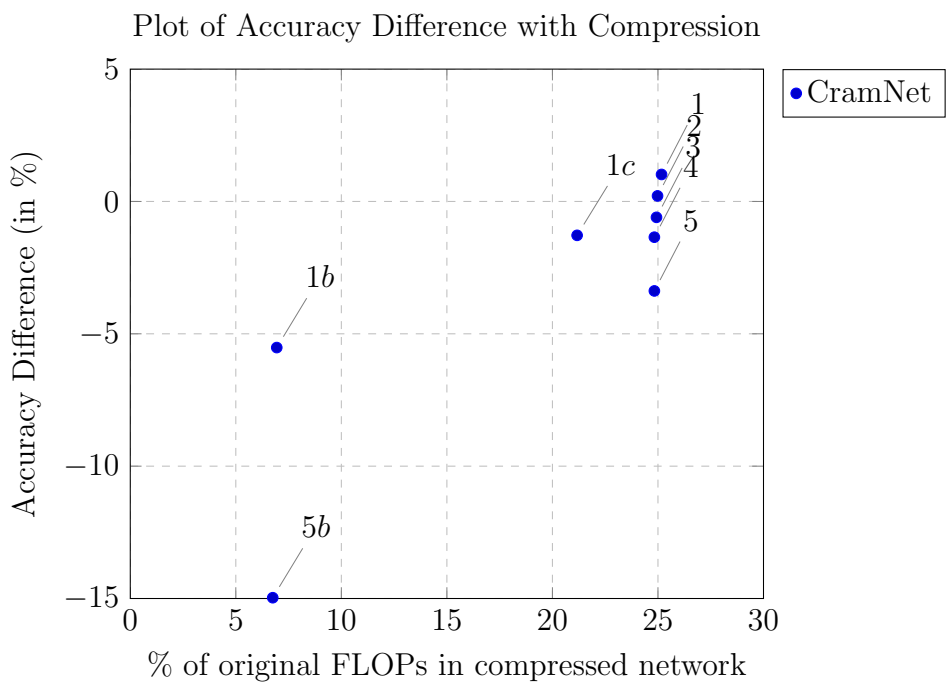

The plot of accuracy vs FLOPs paints a slightly different story.  Despite greatly reducing the parameter count, Tests 1-5 barely alter the number of operations performed.  Interestingly, despite having nearly the same number of operations, Test 1b has approximately a 10\% accuracy improvement over Test 5b.  This indicates that a balance approach between parameter count and floating point operations can produce results that maximize the compression without sacrificing accuracy.

\TUsubsection{CIFAR-10 Comparison}

\begin{figure}[ht]
\centering
\resizebox {1\textwidth} {!} {
\begin{tikzpicture}
\begin{axis}[
    title={Plot of Accuracy Difference with Compression},
    xlabel={\% of original parameters in compressed network},
    ylabel={Accuracy Difference (in \%)},
    xmin=0, xmax=100,
    ymin=-20, ymax=5,
    xtick={0,10,20,30,40,50,60,70,80,90,100},
    ytick={-20,-15,-10,-5,0,5},
    legend pos=outer north east,
    xmajorgrids=true,
    ymajorgrids=true,
    grid style=dashed,
    ]
\addplot+[
    only marks,
    ]
    coordinates {
    (10.26, 1.02)
    (7.28, 0.21)
    (5.80, -0.6)
    (5.47, -1.28)
    (4.85, -5.52)
    (4.31, -1.35)
    (2.82, -3.38)
    (1.10, -14.97)
    };
\addlegendentry{CramNet}
\addplot+[
    only marks,
    ]
    coordinates {
    (68.51, -1)
    };
\addlegendentry{Channel Pruning \cite{CPA}}
\addplot+[
    only marks,
    ]
    coordinates {
    (90, -0.7)
    (70, -0.8)
    (60, -1.2)
    (50, -1.25)
    (30, -4.7)
    };
\addlegendentry{AutoML Model Compression \cite{AutoML}}
\addplot+[
    only marks,
    ]
    coordinates {
    (27.78, 1.43)
    (17.79, 0.92)
    (9.58, 0.88)
    (2.77, -1.17)
    };
\addlegendentry{FitNets \cite{FitNets}}
\addplot+[
    only marks,
    ]
    coordinates {
    (9, -0.19)
    };
\addlegendentry{Layer Wise OBS \cite{L-W-OBS}}
\addplot+[only marks,]
    coordinates {
    (9, -0.79)
    };
\addlegendentry{Learning Both Weights and Connections \cite{LBWC}}
\addplot+[
    only marks,
    ]
    coordinates {
    (89.8, 0.43)
    (57.7, 0.26)
    (51.3, -0.42)
    };
\addlegendentry{Shallowing Deep Networks \cite{Shallow}}
\end{axis}
\end{tikzpicture}
}

\caption{Compression Results for the CIFAR-10 dataset}
\label{fig:CIFAR-10_Results_All}
\end{figure}
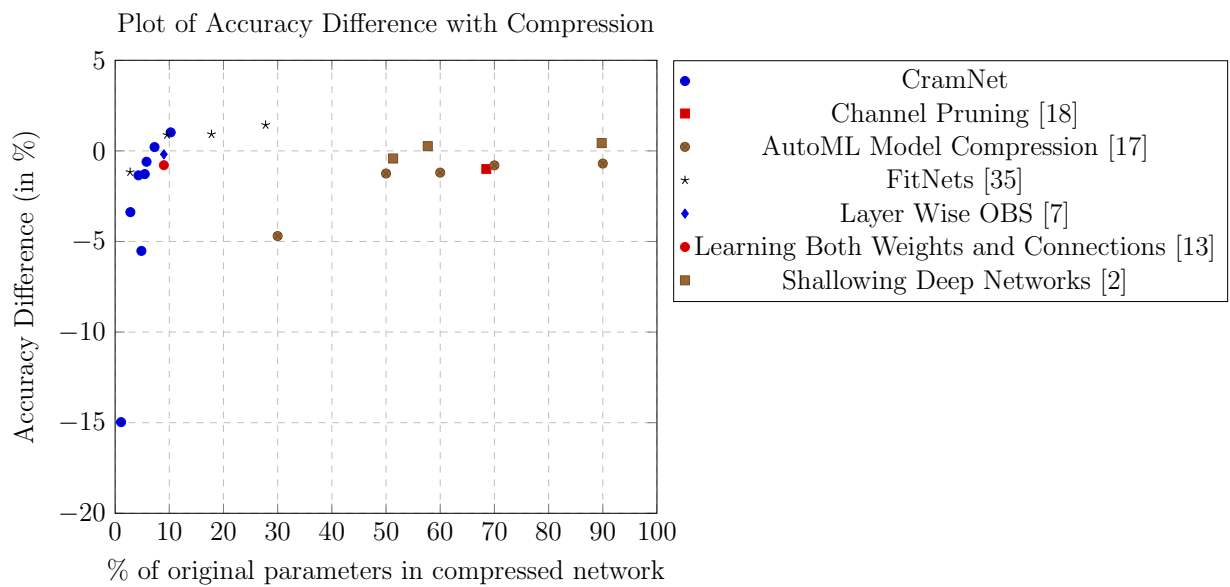

While CIFAR-10 is a very standardized dataset, there is no standard network.  As stated before, each compression algorithm starts with a different initial network.  Using the normalized metrics described in \ref{sec:Metrics} is what makes comparing these methods possible.

Figure \ref{fig:CIFAR-10_Results_All} includes the results from CramNet and six other methods.  The results seem to form two clusters.  First are several channel pruning methods, which use a variety of methods to select which channels to prune, from reinforcement learning \cite{AutoML}, to different channel scores~\cite{Shallow,CPA}.  The second cluster encompasses the sparse pruning and retraining based methods, including CramNet.  The methods in the second cluster shows a marked improvement in compression ratio before losing acccuracy when compared with those in the first cluster.

It is also important to note another difference between the clusters of methods.  Where methods in the first cluster use versions of ResNet or VGG which contain more than 24 million or 100 million parameters respectively, the methods in second cluster use much smaller networks.  FitNets' teacher network starts with 9 million parameters, and CramNet is 1.25 million.  The network used with Layer-wise OBS is unknown, but it contains only 3 convolution layers and 2 fully connected layers.

The result of this disparity in absolute parameter counts is that Test 1 of CramNet contains between 0.5\% and 0.009\% of the parameters used in  the first cluster.  Whether the size of the networks is causative, or a reflection of the effectiveness of the second cluster's methods needs to be investigated in future work.

What is clear is that CramNet stands on the same level as the more effective methods.  Unlike Layer-wise OBS or Learning both Weights and Connections, CramNet produces a dense network.  Unlike FitNets, CramNet places no restrictions on the intermediate layers of the architecture.

\TUsubsection{ImageNet Results}

\begin{table}[ht]
\centering
\begin{tabular}{|c|c|c|c|}
    \hline
    Test \# & \# of Parameters & \% of Network Parameters & Accuracy Difference (\(\pm\%\))  \\ \hline
    \(\emptyset\) & 138,357,544 & 100 & +0 (Baseline) \\ \hline
    1a & 122,699,560 & 88.68 & +2.68 \\ \hline 
    1b & 29,110,568 & 21.04 & -4.59 \\ \hline 
    2a & 120,089,896 & 86.80 & +1.74 \\ \hline 
    2b & 21,782,056 & 15.74 & -9.28 \\ \hline 
    2c & 16,293,096 & 11.78 & -14.47 (Not Yet Complete) \\ \hline 
\end{tabular}
\label{table:ImageNet}
\caption{CramNet Results for the ImageNet dataset}
\end{table}

Test 1a (\ref{B:Test 1a}) and 1b (\ref{B:Test 1b}) begin by leaving the convolutional layers at full depth and restricting the size of the fully connected layers.  Test 1a reduces the last non-output layer to one quarter size, and test 1b reduces both non-output fully connected layers to one quarter size.  Since almost 90\% of the network's parameters are in the fully connected layers, these reductions make a large difference.

Test 2a (\ref{B:Test 2a}), 2b (\ref{B:Test 2b}), and 2c (\ref{B:Test2c}),  bring the compression of the fully connected layers to one eighth.  In addition, Test 2c reduces the last convolutional layer to three eights of the original size.
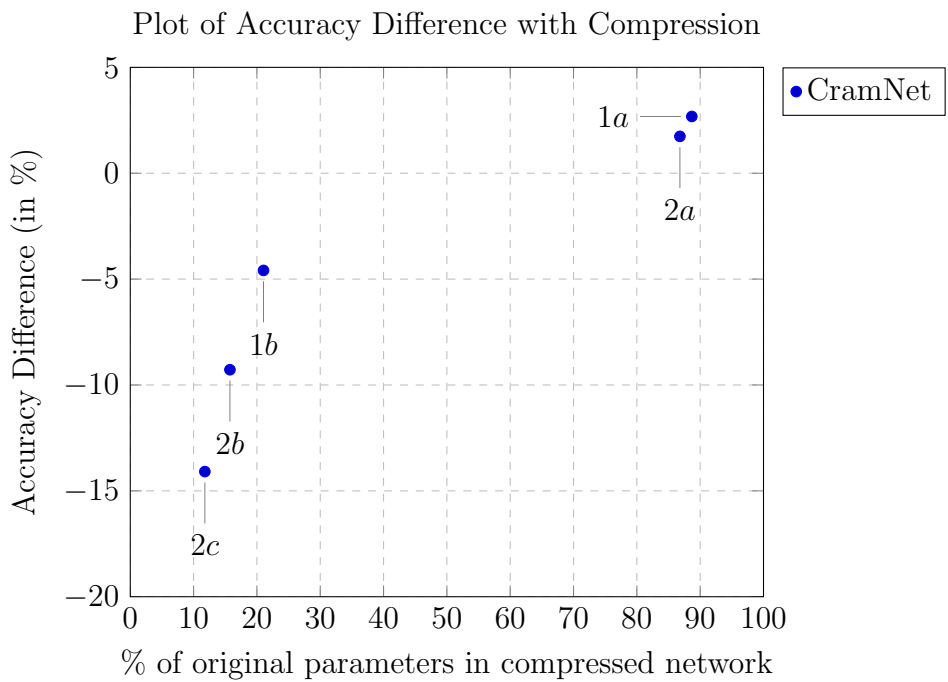
\begin{figure}[ht]
\centering
\begin{tikzpicture}
\begin{axis}[
    title={Plot of Accuracy Difference with Compression},
    xlabel={\% of original parameters in compressed network},
    ylabel={Accuracy Difference (in \%)},
    xmin=0, xmax=100,
    ymin=-20, ymax=5,
    xtick={0,10,20,30,40,50,60,70,80,90,100},
    ytick={-20,-15,-10,-5,0,5},
    legend pos=outer north east,
    xmajorgrids=true,
    ymajorgrids=true,
    grid style=dashed,
    ]
\addplot+[only marks,]
    coordinates {
    (88.68, 2.68)
    (21.04, -4.59)
    (86.80, 1.74)
    (15.74, -9.28)
    (11.78, -14.09) 
    };
\addlegendentry{CramNet}
\addplot[mark=none] coordinates {(88.68, 2.68)} node[pin=180:{$1a$}]{} ;
\addplot[mark=none] coordinates {(21.04, -4.59)} node[pin=270:{$1b$}]{} ;
\addplot[mark=none] coordinates {(86.80, 1.74)} node[pin=270:{$2a$}]{} ;
\addplot[mark=none] coordinates {(15.74, -9.28)} node[pin=270:{$2b$}]{} ;
\addplot[mark=none] coordinates {(11.78, -14.09)} node[pin=270:{$2c$}]{} ;
\end{axis}
\end{tikzpicture}
\caption{CramNet Results for the ImageNet dataset using \% Parameters metric}
\label{fig:ImageNet_Results_Param}
\end{figure}

While further testing would be preferable, hardware restrictions greatly reduce the speed at which the retraining can be performed.  The speed at which data samples can be read from the hard-drive is the limiting factor.  A compressed version of the inputs to the fully connected layers was created to speed up testing, reducing the total time to perform one test to approximately 1 day, most of which was spent reading data from disk.  Test 2c, which includes a convolution layer, and so cannot use the compressed inputs, is estimated to require 3 days.

When using the compressed input, the convolutional layers cannot be retrained.  One possible effect of this is seen during fine tuning.  During fine tuning, the network ``takes in slack'' and reduces under-fitting. Notice how Tests 1 and 2, which have an untouched layer to fine tune produce positive accuracy changes, whereas neither 1b and 2b recover their full values.

\TUsubsection{ImageNet Comparison}

\begin{figure}[ht]
\centering
\resizebox {1\textwidth} {!} {
\begin{tikzpicture}
\begin{axis}[
    title={Plot of Accuracy Difference with Compression},
    xlabel={\% of original parameters in compressed network},
    ylabel={Accuracy Difference (in \%)},
    xmin=0, xmax=100,
    ymin=-20, ymax=5,
    xtick={0,10,20,30,40,50,60,70,80,90,100},
    ytick={-20,-15,-10,-5,0,5},
    legend pos=outer north east,
    xmajorgrids=true,
    ymajorgrids=true,
    grid style=dashed,
    ]
\addplot+[only marks,]
    coordinates {
    (88.68, 2.68)
    (21.04, -4.59)
    (86.80, 1.74)
    (15.74, -9.28)
    (11.78, -14.47) 
    };
\addlegendentry{CramNet}
\addplot+[only marks,]
    coordinates {
    (7.5, -0.77)
    };
\addlegendentry{Learning Both Weights and Connections \cite{LBWC}}
\addplot+[only marks,]
    coordinates {
    (22.05, 0.5)
    (6.87, -4.11)
    (2.67, -16.81)
    };
\addlegendentry{Holistic CNN compression \cite{Holistic}} 
\addplot+[only marks,]
    coordinates {
    (7.5, -0.36)
    };
\addlegendentry{Layer-wise OBS \cite{L-W-OBS}}
\addplot+[only marks,color=magenta,]
    coordinates {
    (7.5, -0.36)
    };
\addlegendentry{DNS \cite{DNS}}
\addplot+[only marks,]
    coordinates {
    (95.01, -1.76)
    (94.33, -2.39)
    (27.65, -3.81)
    (3.24, -8.53)
    (1.84, -14.09)
    };
\addlegendentry{ThiNet \cite{ThiNet}}
\end{axis}
\end{tikzpicture}
}
\caption{Compression Results for the ImageNet dataset}
\label{fig:ImageNet_Results_All}
\end{figure}
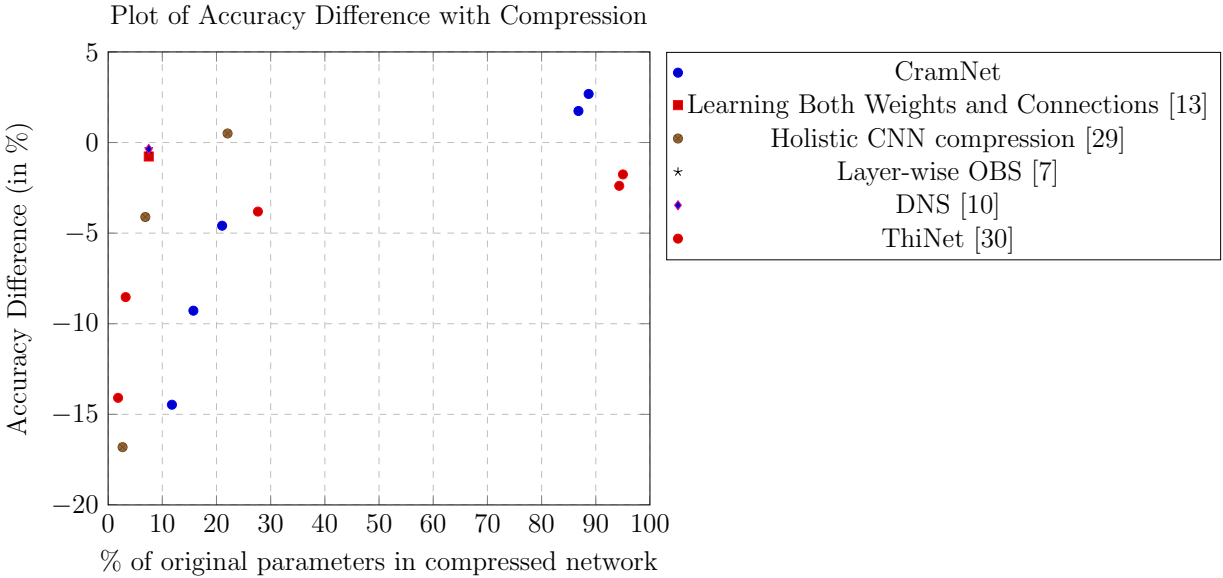

Due to the size of the dataset and the VGG-16 network it is unsurprising that most papers include only a few examples.  Several only provide a single data point.  ThiNet, while it has 5 data points, spreads them out over several variants of the VGG structure.

Clearly, the methods that provide the most compression with the least accuracy loss are Layer-wise OBS, DNS, and LBWC, which are all sparse pruning methods. A close second is a method called Holistic CNN compression~\cite{Holistic}.  This method works in many ways like CramNet, including the loss function and layer-wise retraining, but instead of reducing the number of convolutional filters, it decomposes the filters into multiple one-dimensional filters.  

ThiNet has some disadvantages.  Especially when lightly compressing the network, it loses more accuracy than CramNet.  While it appears better when heavily compressing, this is deceptive.  Part-way through the experiment, they switch from using fully connected layers to a Global Average Pooling (GAP) layer.  Using a GAP reduces the size of the VGG-16 network by more than 100 million nodes, which greatly increases the relative size of the convolutional layers.  So while the compression rates seem better, the networks using GAP are no more pruned than the less compressed networks.

CramNet's accuracy seems to be struggling in comparison to the Holistic compression and the sparse methods, but it can be argued that it is not as bad as it looks. As mentioned above, the fine tuning used for Tests 1a and 2a succeeded well, possibly because they could fine tune one layer more than they compressed.  Test b and c however, had no buffer, and this may cause the reduced accuracy.

This chapter showed the results of CramNet's experimentation on multiple datasets.  It demonstrated great success on the CIFAR-10 dataset, and mixed results on the ImageNet dataset.  The next chapter will discuss conclusions and possible future work.

\TUchapter{Conclusions and Future Work}
We proposed a new network compression algorithm designed to use a teacher network to produce dense output networks. The CramNet method allows architectural flexibility and is able to compress both from and to all sorts of network architectures.  Experiments show that the CramNet method can compress small networks to less than 10\% of the original size without losing accuracy.  The experiments also empirically demonstrate the capability of CramNet to compress both convolutional and fully connected layers.

There are several direction to explore with CramNet.  One focus would be validating CramNet's effectiveness on very deep networks such as ResNet-100.  Another area of exploration is validating CramNet on other layer types, such as bypass layers or Long Short Term Memory (LSTM) layers.  Layer types such as LSTM are very important in applications like action recognition and voice recognition.

Lastly, while the effectiveness of quantization on pruned networks has been shown in \cite{Deep-Compression}, demonstrating that capability on a CramNet compressed network would produce very small networks, well suited for low-bandwidth mobile applications.

Driving the development of network compression methods is the desire to run slow or large networks in real-time or on smaller, cheaper devices.  CramNet demonstrates the potential for enabling larger, more capable networks than have been possible to date.


\addtocontents{toc}{\protect{\hfill \ }}
\addcontentsline{toc}{section}{\contentsadj NOMENCLATURE}
\bibliographyp
\bibliography{Bibliography}
\bibliographystyle{plain}


\appendixpages       
\TUappendix{CIFAR-10 Architectures}
\label{App:A}

The following tables describe the architectures used with the tests on the CIFAR-10 dataset.  All Conv2D layers include a rectified linear unit (RELU) activation.  All fully connected (fc\#) layers also include a RELU activation.  All prediction layers include a softmax activation (see \ref{softmax} for details).

The ``None'' in the output shape represents the placeholder for the number of inputs.  All operations would be replicated across the inputs, and the network would produce that many output vectors.

\TUsection{Default Architecture}
\label{A:Default}
\begin{singlespace}
\begin{verbatim}

Layer (type)                    Output Shape             Param #   
=================================================================
conv2d_1 (Conv2D)              (None, 32, 32, 32)        896       
conv2d_2 (Conv2D)              (None, 30, 30, 32)        9248      
max_pooling2d_1 (MaxPooling2D) (None, 15, 15, 32)        0         
conv2d_3 (Conv2D)              (None, 15, 15, 64)        18496     
conv2d_4 (Conv2D)              (None, 13, 13, 64)        36928     
max_pooling2d_2 (MaxPooling2D) (None, 6, 6, 64)          0         
flatten_1 (Flatten)            (None, 2304)              0         
fc1 (Dense)                    (None, 512)               1180160   
predictions (Dense)            (None, 10)                5130      
=================================================================
Total params: 1,250,858
\end{verbatim}
\end{singlespace}

\TUsection{Test 1}
\label{A:Test 1}
\begin{singlespace}
\begin{verbatim}
Layer (type)                 Output Shape              Param #   
=================================================================
input_14 (InputLayer)          (None, 32, 32, 3)         0         
conv2d_5 (Conv2D)              (None, 32, 32, 16)        448       
conv2d_6 (Conv2D)              (None, 30, 30, 16)        2320      
max_pooling2d_3 (MaxPooling2D) (None, 15, 15, 16)        0         
conv2d_7 (Conv2D)              (None, 15, 15, 32)        4640      
conv2d_8 (Conv2D)              (None, 13, 13, 32)        9248      
max_pooling2d_4 (MaxPooling2D) (None, 6, 6, 32)          0         
flatten_2 (Flatten)            (None, 1152)              0         
fc1 (Dense)                    (None, 96)                110688    
predictions (Dense)            (None, 10)                970       
=================================================================
Total params: 128,314
\end{verbatim}
\end{singlespace}

\TUsection{Test 1b}
\label{A:Test 1b}
\begin{singlespace}
\begin{verbatim}
Layer (type)                 Output Shape              Param #   
=================================================================
input_14 (InputLayer)          (None, 32, 32, 3)         0         
conv2d_5 (Conv2D)              (None, 32, 32, 8)         224       
conv2d_6 (Conv2D)              (None, 30, 30, 8)         1160      
max_pooling2d_3 (MaxPooling2D) (None, 15, 15, 8)         0         
conv2d_7 (Conv2D)              (None, 15, 15, 16)        2320      
conv2d_8 (Conv2D)              (None, 13, 13, 16)        4624      
max_pooling2d_4 (MaxPooling2D) (None, 6, 6, 16)          0         
flatten_2 (Flatten)            (None, 576)               0         
fc1 (Dense)                    (None, 96)                55392    
predictions (Dense)            (None, 10)                970       
=================================================================
Total params: 60,658
\end{verbatim}
\end{singlespace}

\TUsection{Test 1c}
\label{A:Test 1c}
\begin{singlespace}
\begin{verbatim}
Layer (type)                 Output Shape              Param #   
=================================================================
input_14 (InputLayer)          (None, 32, 32, 3)         0         
conv2d_5 (Conv2D)              (None, 32, 32, 16)        448       
conv2d_6 (Conv2D)              (None, 30, 30, 16)        2320      
max_pooling2d_3 (MaxPooling2D) (None, 15, 15, 16)        0         
conv2d_7 (Conv2D)              (None, 15, 15, 32)        4640      
conv2d_8 (Conv2D)              (None, 13, 13, 16)        4624      
max_pooling2d_4 (MaxPooling2D) (None, 6, 6, 16)          0         
flatten_2 (Flatten)            (None, 1152)              0         
fc1 (Dense)                    (None, 96)                55392    
predictions (Dense)            (None, 10)                970       
=================================================================
Total params: 68,394
\end{verbatim}
\end{singlespace}

\TUsection{Test 2}
\label{A:Test 2}
\begin{singlespace}
\begin{verbatim}
Layer (type)                 Output Shape              Param #   
=================================================================
input_14 (InputLayer)          (None, 32, 32, 3)         0         
conv2d_5 (Conv2D)              (None, 32, 32, 16)        448       
conv2d_6 (Conv2D)              (None, 30, 30, 16)        2320      
max_pooling2d_3 (MaxPooling2D) (None, 15, 15, 16)        0         
conv2d_7 (Conv2D)              (None, 15, 15, 32)        4640      
conv2d_8 (Conv2D)              (None, 13, 13, 32)        9248      
max_pooling2d_4 (MaxPooling2D) (None, 6, 6, 32)          0         
flatten_2 (Flatten)            (None, 1152)              0         
fc1 (Dense)                    (None, 64)                73792    
predictions (Dense)            (None, 10)                650       
=================================================================
Total params: 91,098
\end{verbatim}
\end{singlespace}

\TUsection{Test 3}
\label{A:Test 3}
\begin{singlespace}
\begin{verbatim}
Layer (type)                 Output Shape              Param #   
=================================================================
input_14 (InputLayer)          (None, 32, 32, 3)         0         
conv2d_5 (Conv2D)              (None, 32, 32, 16)        448       
conv2d_6 (Conv2D)              (None, 30, 30, 16)        2320      
max_pooling2d_3 (MaxPooling2D) (None, 15, 15, 16)        0         
conv2d_7 (Conv2D)              (None, 15, 15, 32)        4640      
conv2d_8 (Conv2D)              (None, 13, 13, 32)        9248      
max_pooling2d_4 (MaxPooling2D) (None, 6, 6, 32)          0         
flatten_2 (Flatten)            (None, 1152)              0         
fc1 (Dense)                    (None, 48)                55344    
predictions (Dense)            (None, 10)                490   
=================================================================
Total params: 72,490
\end{verbatim}
\end{singlespace}

\TUsection{Test 4}
\label{A:Test 4}
\begin{singlespace}
\begin{verbatim}
Layer (type)                 Output Shape              Param #   
=================================================================
input_14 (InputLayer)          (None, 32, 32, 3)         0         
conv2d_5 (Conv2D)              (None, 32, 32, 16)        448       
conv2d_6 (Conv2D)              (None, 30, 30, 16)        2320      
max_pooling2d_3 (MaxPooling2D) (None, 15, 15, 16)        0         
conv2d_7 (Conv2D)              (None, 15, 15, 32)        4640      
conv2d_8 (Conv2D)              (None, 13, 13, 32)        9248      
max_pooling2d_4 (MaxPooling2D) (None, 6, 6, 32)          0         
flatten_2 (Flatten)            (None, 1152)              0         
fc1 (Dense)                    (None, 32)                36896    
predictions (Dense)            (None, 10)                330       
=================================================================
Total params: 53,882
\end{verbatim}
\end{singlespace}

\TUsection{Test 5}
\label{A:Test 5}
\begin{singlespace}
\begin{verbatim}
Layer (type)                 Output Shape              Param #   
=================================================================
input_14 (InputLayer)          (None, 32, 32, 3)         0         
conv2d_5 (Conv2D)              (None, 32, 32, 16)        448       
conv2d_6 (Conv2D)              (None, 30, 30, 16)        2320      
max_pooling2d_3 (MaxPooling2D) (None, 15, 15, 16)        0         
conv2d_7 (Conv2D)              (None, 15, 15, 32)        4640      
conv2d_8 (Conv2D)              (None, 13, 13, 32)        9248      
max_pooling2d_4 (MaxPooling2D) (None, 6, 6, 32)          0         
flatten_2 (Flatten)            (None, 1152)              0         
fc1 (Dense)                    (None, 16)                18448    
predictions (Dense)            (None, 10)                170   
=================================================================
Total params: 35,274
\end{verbatim}
\end{singlespace}

\TUsection{Test 5b}
\label{A:Test 5b}
\begin{singlespace}
\begin{verbatim}
Layer (type)                 Output Shape              Param #   
=================================================================
input_14 (InputLayer)          (None, 32, 32, 3)         0         
conv2d_5 (Conv2D)              (None, 32, 32, 8)         224       
conv2d_6 (Conv2D)              (None, 30, 30, 8)         1160      
max_pooling2d_3 (MaxPooling2D) (None, 15, 15, 8)         0         
conv2d_7 (Conv2D)              (None, 15, 15, 16)        2320      
conv2d_8 (Conv2D)              (None, 13, 13, 16)        4624      
max_pooling2d_4 (MaxPooling2D) (None, 6, 6, 16)          0         
flatten_2 (Flatten)            (None, 576)               0         
fc1 (Dense)                    (None, 16)                9232    
predictions (Dense)            (None, 10)                170       
=================================================================
Total params: 13,698
\end{verbatim}
\end{singlespace}

\TUappendix{VGG-16 Architectures}
\label{App:B}

The following tables describe the architectures used with the tests on the ImageNet dataset.  All Conv2D layers include a rectified linear unit (RELU) activation.  All fully connected (fc\#) layers also include a RELU activation.  All prediction layers include a softmax activation (see \ref{softmax} for details).

The ``None'' in the output shape represents the placeholder for the number of inputs.  All operations would be replicated across the inputs, and the network would produce that many output vectors.

\TUsection{Default}
\label{B:Default}
\begin{singlespace}
\begin{verbatim}
Layer (type)                 Output Shape              Param #   
=================================================================
input_1 (InputLayer)         (None, 224, 224, 3)       0         
block1_conv1 (Conv2D)        (None, 224, 224, 64)      1792      
block1_conv2 (Conv2D)        (None, 224, 224, 64)      36928     
block1_pool (MaxPooling2D)   (None, 112, 112, 64)      0         
block2_conv1 (Conv2D)        (None, 112, 112, 128)     73856     
block2_conv2 (Conv2D)        (None, 112, 112, 128)     147584    
block2_pool (MaxPooling2D)   (None, 56, 56, 128)       0         
block3_conv1 (Conv2D)        (None, 56, 56, 256)       295168    
block3_conv2 (Conv2D)        (None, 56, 56, 256)       590080    
block3_conv3 (Conv2D)        (None, 56, 56, 256)       590080    
block3_pool (MaxPooling2D)   (None, 28, 28, 256)       0         
block4_conv1 (Conv2D)        (None, 28, 28, 512)       1180160   
block4_conv2 (Conv2D)        (None, 28, 28, 512)       2359808   
block4_conv3 (Conv2D)        (None, 28, 28, 512)       2359808   
block4_pool (MaxPooling2D)   (None, 14, 14, 512)       0         
block5_conv1 (Conv2D)        (None, 14, 14, 512)       2359808   
block5_conv2 (Conv2D)        (None, 14, 14, 512)       2359808   
block5_conv3 (Conv2D)        (None, 14, 14, 512)       2359808   
block5_pool (MaxPooling2D)   (None, 7, 7, 512)         0         
flatten (Flatten)            (None, 25088)             0         
fc1 (Dense)                  (None, 4096)              102764544 
fc2 (Dense)                  (None, 4096)              16781312  
predictions (Dense)          (None, 1000)              4097000   
=================================================================
Total params: 138,357,544
\end{verbatim}
\end{singlespace}

\TUsection{Test 1a}
\label{B:Test 1a}
\begin{singlespace}
\begin{verbatim}
Layer (type)                 Output Shape              Param #   
=================================================================
input_1 (InputLayer)         (None, 224, 224, 3)       0         
block1_conv1 (Conv2D)        (None, 224, 224, 64)      1792      
block1_conv2 (Conv2D)        (None, 224, 224, 64)      36928     
block1_pool (MaxPooling2D)   (None, 112, 112, 64)      0         
block2_conv1 (Conv2D)        (None, 112, 112, 128)     73856     
block2_conv2 (Conv2D)        (None, 112, 112, 128)     147584    
block2_pool (MaxPooling2D)   (None, 56, 56, 128)       0         
block3_conv1 (Conv2D)        (None, 56, 56, 256)       295168    
block3_conv2 (Conv2D)        (None, 56, 56, 256)       590080    
block3_conv3 (Conv2D)        (None, 56, 56, 256)       590080    
block3_pool (MaxPooling2D)   (None, 28, 28, 256)       0         
block4_conv1 (Conv2D)        (None, 28, 28, 512)       1180160   
block4_conv2 (Conv2D)        (None, 28, 28, 512)       2359808   
block4_conv3 (Conv2D)        (None, 28, 28, 512)       2359808   
block4_pool (MaxPooling2D)   (None, 14, 14, 512)       0         
block5_conv1 (Conv2D)        (None, 14, 14, 512)       2359808   
block5_conv2 (Conv2D)        (None, 14, 14, 512)       2359808   
block5_conv3 (Conv2D)        (None, 14, 14, 512)       2359808   
block5_pool (MaxPooling2D)   (None, 7, 7, 512)         0         
flatten (Flatten)            (None, 25088)             0         
fc1 (Dense)                  (None, 4096)              102764544 
fc2 (Dense)                  (None, 1024)              4195328  
predictions (Dense)          (None, 1000)              1025000   
=================================================================
Total params: 122,699,560
\end{verbatim}
\end{singlespace}

\TUsection{Test 1b}
\label{B:Test 1b}
\begin{singlespace}
\begin{verbatim}
Layer (type)                 Output Shape              Param #   
=================================================================
input_1 (InputLayer)         (None, 224, 224, 3)       0         
block1_conv1 (Conv2D)        (None, 224, 224, 64)      1792      
block1_conv2 (Conv2D)        (None, 224, 224, 64)      36928     
block1_pool (MaxPooling2D)   (None, 112, 112, 64)      0         
block2_conv1 (Conv2D)        (None, 112, 112, 128)     73856     
block2_conv2 (Conv2D)        (None, 112, 112, 128)     147584    
block2_pool (MaxPooling2D)   (None, 56, 56, 128)       0         
block3_conv1 (Conv2D)        (None, 56, 56, 256)       295168    
block3_conv2 (Conv2D)        (None, 56, 56, 256)       590080    
block3_conv3 (Conv2D)        (None, 56, 56, 256)       590080    
block3_pool (MaxPooling2D)   (None, 28, 28, 256)       0         
block4_conv1 (Conv2D)        (None, 28, 28, 512)       1180160   
block4_conv2 (Conv2D)        (None, 28, 28, 512)       2359808   
block4_conv3 (Conv2D)        (None, 28, 28, 512)       2359808   
block4_pool (MaxPooling2D)   (None, 14, 14, 512)       0         
block5_conv1 (Conv2D)        (None, 14, 14, 512)       2359808   
block5_conv2 (Conv2D)        (None, 14, 14, 512)       2359808   
block5_conv3 (Conv2D)        (None, 14, 14, 512)       2359808   
block5_pool (MaxPooling2D)   (None, 7, 7, 512)         0         
flatten (Flatten)            (None, 25088)             0         
fc1 (Dense)                  (None, 1024)              25691136 
fc2 (Dense)                  (None, 1024)              1049600  
predictions (Dense)          (None, 1000)              1025000   
=================================================================
Total params: 29,110,568
\end{verbatim}
\end{singlespace}

\TUsection{Test 2a}
\label{B:Test 2a}
\begin{singlespace}
\begin{verbatim}
Layer (type)                 Output Shape              Param #   
=================================================================
input_1 (InputLayer)         (None, 224, 224, 3)       0         
block1_conv1 (Conv2D)        (None, 224, 224, 64)      1792      
block1_conv2 (Conv2D)        (None, 224, 224, 64)      36928     
block1_pool (MaxPooling2D)   (None, 112, 112, 64)      0         
block2_conv1 (Conv2D)        (None, 112, 112, 128)     73856     
block2_conv2 (Conv2D)        (None, 112, 112, 128)     147584    
block2_pool (MaxPooling2D)   (None, 56, 56, 128)       0         
block3_conv1 (Conv2D)        (None, 56, 56, 256)       295168    
block3_conv2 (Conv2D)        (None, 56, 56, 256)       590080    
block3_conv3 (Conv2D)        (None, 56, 56, 256)       590080    
block3_pool (MaxPooling2D)   (None, 28, 28, 256)       0         
block4_conv1 (Conv2D)        (None, 28, 28, 512)       1180160   
block4_conv2 (Conv2D)        (None, 28, 28, 512)       2359808   
block4_conv3 (Conv2D)        (None, 28, 28, 512)       2359808   
block4_pool (MaxPooling2D)   (None, 14, 14, 512)       0         
block5_conv1 (Conv2D)        (None, 14, 14, 512)       2359808   
block5_conv2 (Conv2D)        (None, 14, 14, 512)       2359808   
block5_conv3 (Conv2D)        (None, 14, 14, 512)       2359808   
block5_pool (MaxPooling2D)   (None, 7, 7, 512)         0         
flatten (Flatten)            (None, 25088)             0         
fc1 (Dense)                  (None, 4096)              102764544 
fc2 (Dense)                  (None, 512)               2097664  
predictions (Dense)          (None, 1000)              513000   
=================================================================
Total params: 120,089,896
\end{verbatim}
\end{singlespace}

\TUsection{Test 2b}
\label{B:Test 2b}
\begin{singlespace}
\begin{verbatim}
Layer (type)                 Output Shape              Param #   
=================================================================
input_1 (InputLayer)         (None, 224, 224, 3)       0         
block1_conv1 (Conv2D)        (None, 224, 224, 64)      1792      
block1_conv2 (Conv2D)        (None, 224, 224, 64)      36928     
block1_pool (MaxPooling2D)   (None, 112, 112, 64)      0         
block2_conv1 (Conv2D)        (None, 112, 112, 128)     73856     
block2_conv2 (Conv2D)        (None, 112, 112, 128)     147584    
block2_pool (MaxPooling2D)   (None, 56, 56, 128)       0         
block3_conv1 (Conv2D)        (None, 56, 56, 256)       295168    
block3_conv2 (Conv2D)        (None, 56, 56, 256)       590080    
block3_conv3 (Conv2D)        (None, 56, 56, 256)       590080    
block3_pool (MaxPooling2D)   (None, 28, 28, 256)       0         
block4_conv1 (Conv2D)        (None, 28, 28, 512)       1180160   
block4_conv2 (Conv2D)        (None, 28, 28, 512)       2359808   
block4_conv3 (Conv2D)        (None, 28, 28, 512)       2359808   
block4_pool (MaxPooling2D)   (None, 14, 14, 512)       0         
block5_conv1 (Conv2D)        (None, 14, 14, 512)       2359808   
block5_conv2 (Conv2D)        (None, 14, 14, 512)       2359808   
block5_conv3 (Conv2D)        (None, 14, 14, 512)       2359808   
block5_pool (MaxPooling2D)   (None, 7, 7, 512)         0         
flatten (Flatten)            (None, 25088)             0         
fc1 (Dense)                  (None, 256)               6422784 
fc2 (Dense)                  (None, 512)               131584  
predictions (Dense)          (None, 1000)              513000   
=================================================================
Total params: 21,782,056
\end{verbatim}
\end{singlespace}

\TUsection{Test 2c}
\label{B:Test 2c}
\begin{singlespace}
\begin{verbatim}
Layer (type)                 Output Shape              Param #   
=================================================================
input_1 (InputLayer)         (None, 224, 224, 3)       0         
block1_conv1 (Conv2D)        (None, 224, 224, 64)      1792      
block1_conv2 (Conv2D)        (None, 224, 224, 64)      36928     
block1_pool (MaxPooling2D)   (None, 112, 112, 64)      0         
block2_conv1 (Conv2D)        (None, 112, 112, 128)     73856     
block2_conv2 (Conv2D)        (None, 112, 112, 128)     147584    
block2_pool (MaxPooling2D)   (None, 56, 56, 128)       0         
block3_conv1 (Conv2D)        (None, 56, 56, 256)       295168    
block3_conv2 (Conv2D)        (None, 56, 56, 256)       590080    
block3_conv3 (Conv2D)        (None, 56, 56, 256)       590080    
block3_pool (MaxPooling2D)   (None, 28, 28, 256)       0         
block4_conv1 (Conv2D)        (None, 28, 28, 512)       1180160   
block4_conv2 (Conv2D)        (None, 28, 28, 512)       2359808   
block4_conv3 (Conv2D)        (None, 28, 28, 512)       2359808   
block4_pool (MaxPooling2D)   (None, 14, 14, 512)       0         
block5_conv1 (Conv2D)        (None, 14, 14, 512)       2359808   
block5_conv2 (Conv2D)        (None, 14, 14, 512)       2359808   
block5_conv3 (Conv2D)        (None, 14, 14, 192)       884928   
block5_pool (MaxPooling2D)   (None, 7, 7, 192)         0         
flatten (Flatten)            (None, 9408)              0         
fc1 (Dense)                  (None, 256)               2408704 
fc2 (Dense)                  (None, 512)               131584  
predictions (Dense)          (None, 1000)              513000   
=================================================================
Total params: 21,782,056
\end{verbatim}
\end{singlespace}
\end{document}